\documentclass[lettersize,journal]{IEEEtran}

\usepackage{amsmath,amsfonts,bm}

\def\eqref#1{equation~\ref{#1}}

\def\1{\bm{1}}

\DeclareMathAlphabet{\mathsfit}{\encodingdefault}{\sfdefault}{m}{sl}
\SetMathAlphabet{\mathsfit}{bold}{\encodingdefault}{\sfdefault}{bx}{n}

\usepackage{amsthm}

\newtheoremstyle{nonitalictheorem}
  {\topsep}         
  {\topsep}         
  {\normalfont}     
  {}                
  {\bfseries}       
  {.}               
  {.5em}            
  {}                

\usepackage{booktabs} 
\usepackage{hyperref}
\usepackage{url}
\usepackage{amsmath}
\usepackage{amsfonts}
\usepackage{latexsym}
\usepackage{enumerate}
\usepackage{wasysym}
\usepackage{lipsum}
\usepackage{subfig}
\usepackage{graphicx}
\usepackage{float}
\usepackage{amssymb}
\usepackage{amsmath}
\usepackage{enumitem}
\usepackage{amsthm}
\newtheorem{theorem}{Theorem}
\newtheorem{assumption}{Assumption}
\newtheorem{lemma}{Lemma}
\newtheorem{definition}{Definition}
\newtheorem{proposition}{Proposition}

\theoremstyle{nonitalictheorem}

\usepackage{wrapfig,commath}
\usepackage{color}
\usepackage{algorithm}
\usepackage{algorithmic}
\allowdisplaybreaks[4]

\ifCLASSOPTIONcompsoc
  \usepackage[nocompress]{cite}
\else
  \usepackage{cite}
\fi
\ifCLASSINFOpdf
\else
\fi
\begin{document}
%
\title{Accelerating Asynchronous Federated Learning Convergence via Opportunistic Mobile Relaying}

\author{Jieming Bian,
    Jie Xu
\IEEEcompsocitemizethanks{\IEEEcompsocthanksitem Jieming Bian and Jie Xu are with the Department
of Electrical and Computer Engineering, University of Miami, Coral Gables,
FL, 33146, USA. Email: {jxb1974, jiexu}@miami.edu.\protect\\
}}

\maketitle

\begin{abstract}
This paper presents a study on asynchronous Federated Learning (FL) in a mobile network setting. The majority of FL algorithms assume that communication between clients and the server is always available, however, this is not the case in many real-world systems. To address this issue, the paper explores the impact of mobility on the convergence performance of asynchronous FL. By exploiting mobility, the study shows that clients can indirectly communicate with the server through another client serving as a relay, creating additional communication opportunities. This enables clients to upload local model updates sooner or receive fresher global models. We propose a new FL algorithm, called FedMobile, that incorporates opportunistic relaying and addresses key questions such as when and how to relay. We prove that FedMobile achieves a convergence rate $O(\frac{1}{\sqrt{NT}})$, where $N$ is the number of clients and $T$ is the number of communication slots, and show that the optimal design involves an interesting trade-off on the best timing of relaying. The paper also presents an extension that considers data manipulation before relaying to reduce the cost and enhance privacy. Experiment results on a synthetic dataset and two real-world datasets verify our theoretical findings.  
\end{abstract}


\maketitle

\section{Introduction}
Federated learning (FL) is a distributed machine learning approach in which numerous clients possessing decentralized data cooperate to develop a shared model, under the guidance of a central server \cite{mcmahan2017communication}. The majority of FL algorithms \cite{mcmahan2017communication, li2019convergence, khaled2020tighter, yang2020achieving, yu2019parallel, chen2020wireless, li2023enhancing, li2023icb, liu2020privacy, zheng2022aggregation} explore the synchronous communication setting, where clients can periodically and concurrently synchronize and interact with the server. In this context, communication frequency is a crucial aspect of the FL algorithm design, as it assumes that the communication channel between clients and the server is consistently and universally accessible. However, this assumption is often unrealistic in real-world systems, where clients may only have intermittent opportunities to communicate with the server and display varying communication patterns. For instance, mobile clients (such as mobile devices or vehicles) can only connect with the server (e.g., base stations, sensing hubs, roadside units) when they come within the server's communication range \cite{guerna2022roadside}. Consequently, in such mobile systems, FL must be conducted asynchronously, conforming to each client's unique communication pattern with the server.

The asynchronous nature of these mobile systems renders the methods developed for synchronous FL ineffective. As a result, asynchronous FL methods are necessary to tackle such systems. However, the body of literature on asynchronous FL is notably smaller compared to its synchronous counterpart. Although some insights have been gained (for example, \cite{avdiukhin2021federated} demonstrates that the convergence rate of asynchronous FL can match that of synchronous FL given the same communication interval), the performance of asynchronous FL is significantly constrained by the arbitrary communication patterns of individual clients, which are not an algorithm parameter. With only sporadic client-server interactions, asynchronous FL may converge slowly or even fail to achieve convergence.

To tackle the problem arising from sporadic client-server interactions, in this paper, we focus on exploiting the communication opportunities among clients within the mobile system, an aspect largely overlooked in previous works. As devices in the mobile system continually move over time, numerous client-client encounters are created. These meetings allow a client to indirectly communicate with the server by using another client as a relay, thereby generating additional communication opportunities (both uploading and downloading) with the server. Specifically, a client's local model updates can be uploaded to the server sooner if the relay client connects with the server before the sending client does. Similarly, the client can receive a more recent global model from a relay client that has recently connected with the server.

However, the added benefits provided by the relay for either uploading or downloading introduce new challenges. Firstly, determining when to upload (download) via relaying and selecting the appropriate relay is crucial. If a client uploads to a relay too early, only minimal new information can be transmitted to the server, as the client has completed just a few local steps. Conversely, if the upload to the relay is delayed, the benefit of utilizing the relay may be significantly diminished. Secondly, determining how to relay the local model updates in a manner that prevents duplication of updates received by the server and maintains storage efficiency is also essential. To address these challenges, we propose a novel asynchronous FL algorithm with opportunistic relaying, called FedMobile. The main contributions of our work are as follows:
\begin{itemize}
\item Our work focuses on harnessing the benefits of utilizing relaying resulting from client-client meetings, an aspect not previously explored in mobile FL. We propose a new asynchronous FL algorithm, FedMobile, which accelerates convergence while maintaining storage efficiency and preventing duplicate update transmissions.
\item We offer a theoretical convergence analysis of FedMobile and unveil an intriguing trade-off regarding the optimal timing of relaying, which can, in theory, enhance the convergence speed.
\item We introduce an extension of FedMobile where clients and their relays transmit data with quantization/compression, reducing the communication costs caused by additional relaying.
\item We perform extensive experiments on a synthetic dataset and two real-world datasets to corroborate our theoretical findings. The results confirm that our proposed method outperforms the state-of-the-art asynchronous FL method by significantly reducing time consumption.
\end{itemize}

The rest of this paper is structured as follows. Section II reviews related works. In Section III, we present the system model and briefly revisit the state-of-the-art asynchronous FL method \cite{avdiukhin2021federated}. FedMobile and its extension are detailed in Section IV, while Section V provides the theoretical convergence analysis. Section VI presents an extension for a more general case. Experimental results are reported in Section VII.Finally, Section VIII concludes the paper. All technical proofs can be found in the Supplementary material.

\section{Related Work}
\subsection{Federated Learning}
FedAvg \cite{mcmahan2017communication} uses local stochastic gradient descent (SGD) to reduce the number of communications between the server and clients. Convergence has been analyzed for FedAvg \cite{li2019convergence, khaled2020tighter, yang2020achieving, yu2019parallel, liu2022right} and its variants (e.g., momentum variants \cite{wang2019slowmo, liu2020accelerating} and variance-reducing variants \cite{konevcny2016federated, karimireddy2020scaffold}) in both iid and non-iid data settings. However, the majority of works on FL study the synchronous setting and treat the communication frequency as a tunable parameter of the algorithm.

In recent years, asynchronous distributed optimization and learning have been extensively studied. Works such as \cite{mitliagkas2016asynchrony, hadjis2016omnivore, chen2016revisiting, zheng2017asynchronous, dai2018toward} focus on single SGD steps by distributed nodes with iid data distributions, which do not represent typical FL settings. The literature on asynchronous FL is smaller and has varied emphases. For instance, some existing works \cite{chen2018lag, smith2017federated, nishio2019client} still assume universal communication at all times, with asynchronicity resulting from an algorithmic decision rather than a constraint. Other works \cite{van2020asynchronous, chai2021fedat, chen2020asynchronous, li2021stragglers} employ asynchronous model aggregation to tackle the "straggler" issue in synchronous FL. The asynchronous setting most similar to ours is examined in \cite{avdiukhin2021federated, basu2019qsparse}, which consider arbitrary communication patterns. However, these studies only focus on the interaction between the server and the clients, overlooking the interactions among clients themselves. In this work, we propose FedMobile, an algorithm that leverages the benefits of relaying through client interactions, and demonstrate its superior convergence performance both theoretically and empirically compared to state-of-the-art asynchronous method \cite{avdiukhin2021federated}.

Our work is remotely related to decentralized FL \cite{lalitha2019peer, zhang2021d2d, xing2020decentralized} (and some hybrid FL works \cite{guo2021hybrid}) where clients can also communicate among themselves during the training process, typically by using a type of gossip algorithm to exchange local model information. However, these works assume a fixed topology of clients and the communication among the clients is still synchronous and periodic. 

\subsection{Opportunistic Relaying}
Opportunistic relaying is a wireless communication technique where intermediate nodes in a wireless network temporarily act as relay nodes to forward data packets to their intended destination. The fundamental concept behind this technique is to harness the unused resources in the network, such as energy, idle time, and bandwidth, to enhance the network's overall performance. Although the idea of opportunistic relaying has been well extended and studied in many fields (e.g. wireless networks \cite{cui2009opportunistic}, UAV\cite{9112744}), to our best knowledge, there is no prior work on opportunistic relaying in FL. Additionally, previous studies concentrate on transmitting the unchanged value in situations where no learning process is performed. Our work offers the first principled investigation of the interplay between opportunistic relaying via client-client communication and the convergence of FL, and how to design the optimal relaying strategy to maximize the convergence speed in the learning process.

\section{Model and Preliminaries}
\label{pre}
We consider a mobile FL system with one server and $N$ mobile clients. The mobile clients work together to \textcolor{black}{optimize the model parameters $x \in \mathbb{R}^d$ by minimizing the global objective function $f(x)$}:
\begin{align}
    \min_x f(x) = \frac{1}{N}\sum_{i=1}^N f_i(x) = \frac{1}{N}\sum_{i=1}^N \mathbb{E}_{\zeta_i}[F_i(x, \zeta_i)]
\end{align}
where $f_i:\mathbb{R}^d \to \mathbb{R}$ is a non-convex loss function for client $i$ and $F_i$ is the estimated loss function based on a mini-batch data sample $\zeta_i$ drawn from client $i$'s own dataset. 

\textbf{Mobility Model}. We consider a discrete time system where time is divided into slots of equal length. Clients move independently in a network and make contact with the server only at certain time slots. At each time $t$ and for each client $i$, let $\tau^\text{last}_i(t)$ be the last time when client $i$ meets the server by $t$ (including $t$), and $\tau^\text{next}_i(t)$ be the next time when client $i$ will meet the server (excluding $t$). We assume that at any given time $t$, client $i$ can \textbf{estimate} $\tau^\text{next}_i(t)$. For the sake of proving theoretical convergence, we will assume in the theoretical analysis section that at any $t$, client $i$ knows its exact $\tau^\text{next}_i(t)$. However, in the experiment section, we will compare performances under both the scenarios where $\tau^\text{next}_i(t)$ is exactly known by the client and where $\tau^\text{next}_i(t)$ is estimated by the client. The communication pattern is arbitrary and different for different clients but we assume that the time interval between any two consecutive server meetings for any client is bounded by $\Delta$, i.e., $\tau^\text{next}_i(t) - \tau^\text{last}_i(t) \leq \Delta, \forall t, \forall i$. 


Clients can also meet among themselves due to mobility. When two clients meet, they can communicate with each other via, e.g., device-to-device (D2D) communication protocols \cite{asadi2014survey}. For ease of analysis, we assume that at each time $t$, a client can meet at most one other client (the extension to multiple clients is straightforward). Let $\rho \in [0, 1]$ be the probability of a client meeting another client at a time slot. When $\rho = 0$, clients do not meet with each other, thus degenerating to the conventional case. We assume that the client-server meetings remain unchanged regardless of the value of $\rho$. In the following, we provide a realistic example of a Vehicular Ad Hoc Network (VANET).

\begin{figure}[h]
	\centering
	\includegraphics[width=0.99\linewidth]{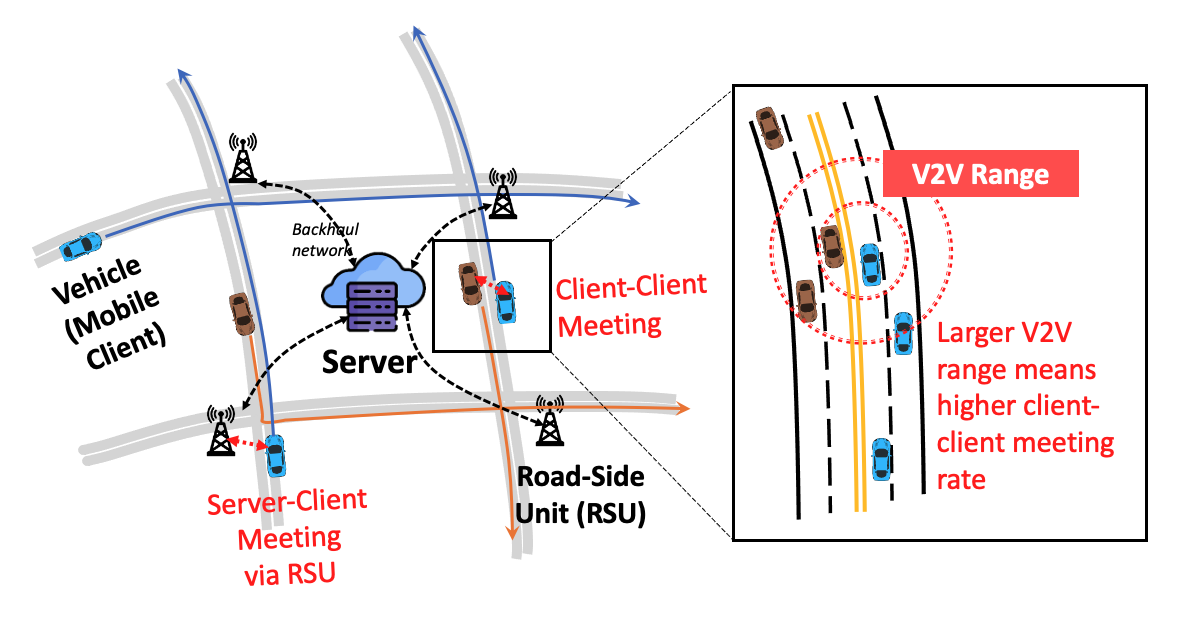}
 \vspace{0.1in}
	\caption{An illustration of Mobile FL system in VANET.} \label{fig:vanet}
\end{figure}

Fig. \ref{fig:vanet} is an illustration of Vehicular Ad Hoc Network (VANET). Consider a road network where many vehicles (as mobile clients) are moving and performing FL (e.g., for traffic prediction) using on-board computing power. Some roadside units (RSUs) are deployed in the road network which the vehicles can connect to via wireless. All pertinent RSUs and the  server are interconnected through high-speed communication technologies such as fiber optic networks \cite{reis2013deploying}. This ensures that the transmission latency between the server and the RSUs is exceedingly low and can, therefore, be disregarded. In a typical deployment, the RSUs are only deployed in certain parts of the road network (e.g., some road intersections) and do not cover the entire road network \cite{guerna2022roadside}. Therefore, the vehicles (clients) can communicate with the server only when they enter the communication range of a RSU. Because vehicles have different speeds and routes, their meeting times with servers (via the RSUs) are naturally asynchronous.

In the VANET use case, vehicles (clients) often have a pre-determined route and move at a roughly constant speed. Thus, it is easy to calculate the next server meeting given the next RSU location and the vehicle trajectory. If a vehicle’s moving speed is fixed, then the client-server meetings do not change (because distances between RSUs cannot be changed). A client-client meeting occurs if two clients enter the Vehicle-to-Vehicle (V2V) communication (e.g., Wi-Fi Direct or LTE Direct) range of each other. Thus, by increasing the communication range (e.g., by using a larger transmission power), more client-client meetings can occur while the client-server meetings stay the same. Thus, $\rho$ is a parameter that models the client-client meeting rates in this case. A smaller client-client communication range results in a smaller $\rho$ while a larger client-client communication range results in a larger $\rho$.

\textbf{State-of-the-art Asynchronous FL method}. 
Before we introduce our proposed method, let's briefly revisit the state-of-the-art asynchronous FL method, ASYNC \cite{avdiukhin2021federated}, which operates under arbitrary communication patterns.

\underline{Local Model Update}. For any client $i$, when it meets the server at $t$, it downloads the current global model $x^t$. The client then uses $x^t_i = x^t$ as the initial model to train a new local model using its local dataset until it meets the server again. This is done by using a mini-batch SGD method:
\begin{align}
    x^{s+1}_i = x^s_i - \eta g^s_i, \forall s = t, ..., \tau^\text{next}_i(t)-1
    \label{localModelUpdate}
\end{align}
where $g^s_i = \nabla F_i(x^s_i, \zeta^s_i)$ is the stochastic gradient on a randomly drawn mini-batch $\zeta^s_i$ and $\eta$ is the learning rate. Here, we assume that a client performs one step SGD at each time slot to keep the notations simple. Let $m^t_i \in \mathbb{R}^d$ be the cumulative local updates (CLU) of client $i$ at time $t$ since its last meeting with the server, which is updated recursively as follows:
\begin{align}
    &m^t_i = \eta g^{t-1}_i,~~\text{if}~~t = \tau^\text{last}_i(t) + 1;\nonumber\\~~~~&
    m^t_i = m^{t-1}_i + \eta g^{t-1}_i,~~\forall t = \tau^\text{last}_i(t) + 2, ..., \tau^\text{next}_i(t)
\end{align}

\underline{Global Model Update}. At any $t$, let $\mathcal{S}^t$ denote the set of clients who meet the server ($\mathcal{S}^t$ may be empty). These clients upload their CLUs to the server which then updates the global model as
\begin{align}
    x^t = x^{t-1} - \frac{1}{N}\sum_{i \in \mathcal{S}^t} m^{t}_i
\end{align}
The updated global model $x^t$ is then downloaded to each client $i$ in $\mathcal{S}^t$, and the client starts its local training with a new initial model $x^t_i = x^t$.

\begin{table}[ht]
\centering
\caption{Key Notations}
\begin{tabular}{cc}
\toprule
\textbf{Symbol} & \textbf{Semantics} \\
\midrule
$t$ & time slot which is measured by one local update\\
$N$ & the number of mobile clients \\
\textcolor{black}{$\mathcal{S}^t$} & \textcolor{black}{the set of clients who meet the server}\\
$f_i$ & non-convex loss function for client $i$ \\
$F_i$ & estimated loss function based on mini-batch data sample \\
\textcolor{black}{$\eta$} & \textcolor{black}{the local learning rate}\\
\textcolor{black}{$g$} & \textcolor{black}{the stochastic gradient}\\
$\tau_i^{last}$ & the last time when client $i$ meets the server\\
$\tau_i^{next}$ & the next time when client $i$ will meet the server\\
$\Delta$ & the upper bound of times interval\\
$\theta, \Theta$ & the upload search interval parameters\\
$\omega, \Omega$ & the download search interval parameters\\
$m_i^t$ & the CLU stored by client $i$ at time $t$\\
$\tilde{m}^t_i$ & the manipulated CLU\\
\textcolor{black}{$n_i^t$} & \textcolor{black}{the CLU only contains client $i$'s own updates} \\
\textcolor{black}{$Q (\cdot)$} & \textcolor{black}{the quantization operator} \\
\textcolor{black}{$\mu_i^t$} & \textcolor{black}{the perturbation noise}\\
\textcolor{black}{$\epsilon_i^t$} & \textcolor{black}{the difference between $\tilde{m}^t_i$ and $m_i^t$}\\

\bottomrule
\end{tabular}
\end{table}

\section{FedMobile}
In the vanilla asynchronous FL described in the Section \ref{pre}, a client can upload its CLU and download the global model  only when it meets the server. When the meeting intervals are long, however, this information cannot be exchanged in a timely manner, thus hindering the global training process. To overcome this issue, we take advantage of \textit{mobility} and propose FedMobile that creates indirect client-server communication opportunities, thereby improving the FL convergence speed. In a nutshell, a client $i$ can use another client $j$ as a relay to upload its CLU to the server (or download the global model from the server) if client $j$'s next (or last) server meeting is earlier (or later) than client $i$'s. Thus, the CLU can be passed to the server sooner (or the client can train based on a fresher global model). We call such client $j$ an upload (or download) \textit{relay} for client $i$.

For now, we assume that a client can only use \textit{at most one} upload relay and \textit{at most one} download relay between any two consecutive server meetings, considering the extra cost incurred due to the client-client communication. However, FedMobile can be easily extended (we discuss it in The General Case section). Next, we describe separately how FedMobile handles uploading and downloading, which are essentially the dual cases to each other. 

\subsection{Uploading CLU via Relaying} \label{upload_main}

\begin{figure}[t]
	\centering
	\includegraphics[width=0.63\linewidth]{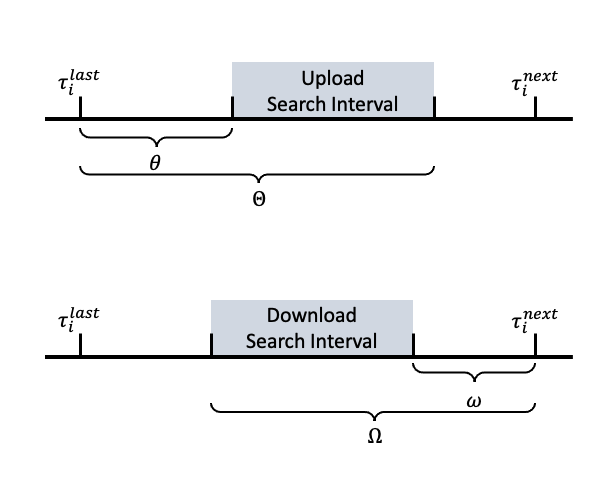}
	\caption{The upload/download search intervals.} \label{fig:search_interval}
\vspace{-0.2in}
\end{figure}

\textcolor{black}{\textbf{Upload Timing and Relay Selection.}} Since a client can use the upload relay at most once between its two consecutive server meetings, the timing of uploading via relaying is crucial. If client $i$ uploads too early (i.e., close to $\tau^\text{last}_i$), then the CLU has little new information since the client has run just a few mini-batch SGD steps. If client $i$ uploads too late (i.e., close to $\tau^\text{next}_i$), then the CLU will be uploaded to the server late even if a relay is used. To make this balance, FedMobile introduces a notion called \textit{upload search interval} defined by two parameters $\theta$ and $\Theta$, where $0 \leq \theta \leq \Theta \leq \Delta$ (See Fig. \ref{fig:search_interval}). Client $i$ will upload its CLU via a relay only during the interval $[\tau^\text{last}_i + \theta, \tau^\text{last}_i +\Theta]$. In addition, not every other client that client $i$ meets during the search interval is qualified. \textcolor{black}{We here define \textit{semi-qualified} upload relay and \textit{qualified} upload relay as follows:}
\begin{definition}
    \textcolor{black}{
    Client $j$ is a \textbf{semi-qualified} upload relay if $\tau^\text{next}_j \leq \tau^\text{last}_i +\Theta$, i.e., client $j$ is able to relay client $i$'s CLU to the server before the end of the search interval.
    Further, Client $j$ is a \textbf{qualified} upload relay if in addition $\tau^\text{next}_j < \tau^\text{next}_i$, i.e., client $j$ can indeed deliver the CLU earlier than client $i$'s own server meeting \footnote{It is possible that $\tau^\text{last}_i +\Theta \geq \tau^\text{next}_i$ due to the fixed value of $\Theta$, so a semi-qualified upload relay is not necessarily qualified. }.}
\end{definition}

FedMobile picks the \textit{first} qualified upload relay during the search interval. Note that it is possible that no qualified upload relay is met, in which case no uploading via relaying is performed. The determination of setting parameters $\theta$ and $\Theta$ will be addressed in Section \ref{section_conv}, following the theoretical analysis of their impact on convergence.

\textcolor{black}{\textbf{Upload Relay Mechanism.}} FedMobile implements a streamlined and storage-efficient mechanism to guarantee the delivery of a specific piece of information to the server precisely once, all while avoiding the need to retain client ID information. When a CLU exchange event occurs at time $t$ involving a sender client $i$ and a relay client $j$, the following steps are employed:

\underline{RESET} (by sender): After sending its current CLU $m^t_i$, sender client $i$ resets its CLU to $m^t_i := 0$


\underline{COMBINE} (by relay): \textcolor{black}{After receiving $m^t_i$ from client $i$, client $j$ updates it stored CLU $m^t_j$ by incorporating $m^t_i$, i.e., $m^t_j := m^t_j + m^t_i$. }

In this way, FedMobile essentially offloads the uploading task of $m^t_i$ from client $i$ to client $j$, who, by our design, has a sooner server meeting time than client $i$. 


\textcolor{black}{\textbf{Remark}: Upon transmission of the local update from the sender (client $i$) to relay (client $j$), the local training processes of both the sender and the relay remain unaffected. Specifically, each continues to train locally based on their local models. This process persists until each client reaches its next scheduled server communication, denoted as $\tau^\text{next}_i(t)$ for client $i$ and $\tau^\text{next}_j(t)$ for client $j$.}

\textbf{Remark}: 
In FedMobile, a client has the capability to function as both a sender and a relay between consecutive server meeting times. When a client $i$ transmits message $m^t_i$ to client $j$, it is possible that client $i$ has already received CLUs (Client Local Updates) from other clients. Consequently, the message $m^t_i$ may contain CLUs associated with those other clients. In our relay mechanism, the relay client is not required to store the ID information of the sender client. \textcolor{black}{Even if the relay client were to gain knowledge of the sender client's ID information, it would still face substantial difficulty in determining which clients' CLUs are included in the CLU sent by the sender client.} Due to the presence of CLUs mixed with unknown clients' local updates, the relay client faces significant difficulty in discerning the personal privacy of the sender client.

\textbf{Remark}: For higher communication efficiency and/or better privacy protection, the clients may send an altered CLU to the relay for uploading. We discuss this extension and provide its convergence analysis in the following subsection. 

\begin{figure}[h]
    \centering
    \begin{minipage}[t]{0.7\linewidth}
	\includegraphics[width=1\linewidth]{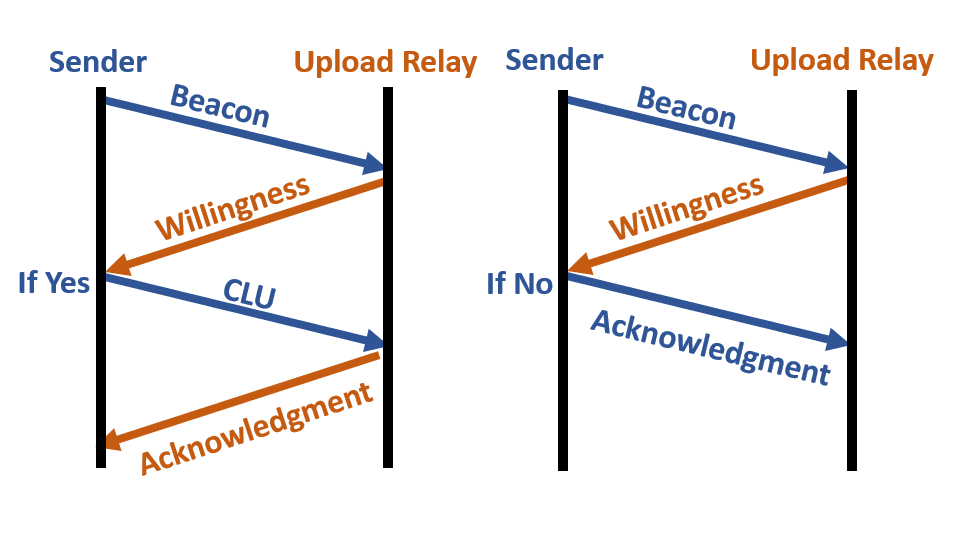}
	\caption{\textcolor{black}{Upload Relay Protocol}} \label{fig:upload}
    \end{minipage}
    \vspace{-10 pt}
\end{figure}

\textcolor{black}{
\textbf{Upload Relay Protocol.}
In the upload relay process, the sender client \(i\) identifies a potential upload relay client \(j\) within the communication range. Client \(i\) initiates the process by sending a `Beacon' message to client \(j\) to inquire about its willingness to serve as an upload relay. If client \(j\) agrees, it responds with a `Willingness' message, including its estimated next meeting time with the server (\(\tau^\text{next}_j\)). Client \(i\) then decides whether to select client \(j\) as its upload relay based on \(\tau^\text{next}_j\) and its own next meeting time (\(\tau^\text{next}_i\)). If client \(j\) is not chosen, client \(i\) sends an `Acknowledgement' message to conclude the interaction. Otherwise, client \(i\) transmits the CLU \(m_i^t\) to client \(j\), who acknowledges receipt and concludes the process. This protocol is depicted in Fig \ref{fig:upload}.
}

\subsection{Downloading Global Model via Relaying} \label{download_subsection}

\textcolor{black}{\textbf{Download Timing and Relay Selection.}} Similar to the uploading CLU case, when downloading a global model via relaying also involves a trade-off. FedMobile introduces a \textit{download search interval} to make this balance, which is defined by two parameters $\omega$ and $\Omega$, where $0 \leq \omega \leq \Omega \leq \Delta$ (see Fig. \ref{fig:search_interval}). Given client $i$'s next server meeting time $\tau^\text{next}_i$, client $i$ will only download a new global model via relaying during the search interval $[\tau^\text{next}_i - \Omega, \tau^\text{next}_i - \omega]$. Similarly, \textcolor{black}{We here define \textit{semi-qualified} download relay and \textit{qualified} download relay as follows:}

\begin{definition}
\textcolor{black}{
    Client $j$ is a \textbf{semi-qualified} download relay if $\tau^\text{last}_j \geq \tau^\text{next}_i - \Omega$, i.e., client $j$'s global model is less than $\Omega$ time slots older than client $i$'s next global model directly from the server. 
    Further client $j$ is a \textbf{qualified} download relay if in addition $\tau^\text{last}_j > \tau^\text{last}_i$, i.e., client $j$'s global model is indeed fresher than client $i$'s current global model that it received directly from the server. }
\end{definition}
FedMobile picks the \textit{first} qualified download relay during the search interval. Again, it is possible that no qualified download relay is met, in which case no downloading via relaying occurs. The determination of setting parameters $\omega$ and $\Omega$ will be addressed in Section \ref{section_conv}, following the theoretical analysis of their impact on convergence.

\textcolor{black}{\textbf{Download Relay Mechanism.}} To be able to relay a global model to other clients, every client keeps a copy of the most recent global model that it received (from either the server or another client). We denote this copy for client $j$ by $x^{\psi_j(t)}$, where $\psi_j(t)$ is the time version (or timestamp) of the global model. Upon a global model exchange at time $t$ between a receiver client $i$ and a relay client $j$:

\underline{REPLACE} (by receiver): After receiving $x^{\psi_j(t)}$ from client $j$, client $i$ replaces its local model with $x^{\psi_i(t)}$, i.e., $x^t_i := x^{\psi_j(t)}$, and resumes the local training steps. 

\textbf{Remark}: Client $i$ also replaces its global model copy with $x^{\psi_j(t)}$ since it is a fresher version by our design. Thus, $\psi_i(t)$ is updated to $\psi_j(t)$. 

\textbf{Remark}: An alternative to the current downloading scheme is that relay client $j$ simply sends its current local model $x^t_{j}$ to client $i$, who then replaces its current local model $x^t_i$ with $x^t_j$, i.e., $x^t_i:= x^t_j$. In this way, the clients do not have to keep a copy of the most recent global model, thereby reducing the stored data. The convergence analysis is not affected by this change and the same convergence bound can be proved (see the proof of Theorem \ref{thm:convergence}). 

\begin{figure}[h]
    \centering
    \begin{minipage}[t]{0.7\linewidth}
	\includegraphics[width=1\linewidth]{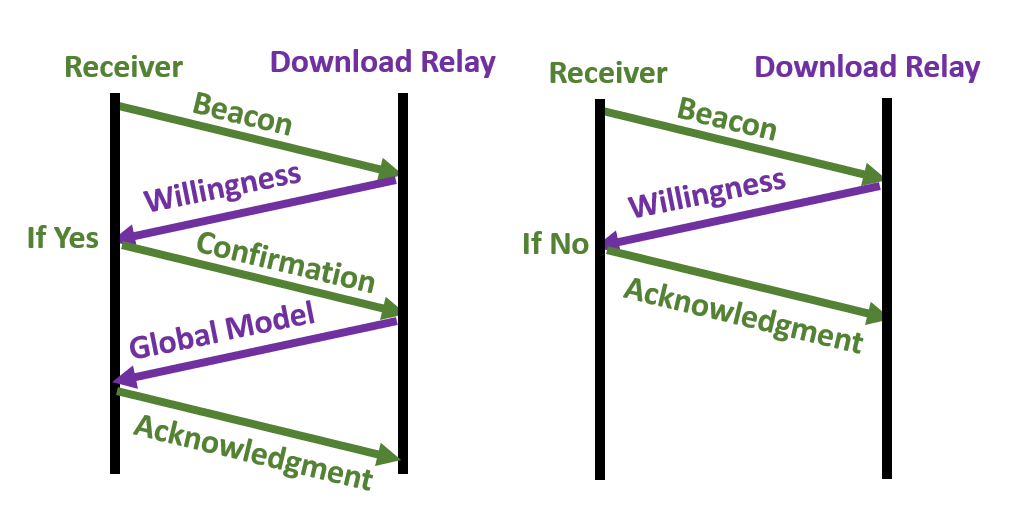}
	\caption{\textcolor{black}{Download Relay Protocol}} \label{fig:download}
    \end{minipage}
    \vspace{-10 pt}
\end{figure}

\textcolor{black}{
\textbf{Download Relay Protocol.}
In the download relay scenario, the receiver (client \(i\)) similarly sends a `Beacon' message to a potential download relay (client \(j\)). Upon client \(j\)'s agreement, indicated by a `Willingness' message containing its last server meeting time (\(\tau^\text{last}_j\)), client \(i\) compares this with its own last meeting time (\(\tau^\text{last}_i\)). If client \(j\) is selected as the relay, client \(i\) sends a `Confirmation' message, prompting client \(j\) to transmit its stored global model version \(x^{\psi_j(t)}\) to client \(i\), who then acknowledges receipt. If client \(j\) is not chosen, client \(i\) directly sends an `Acknowledgement' message to conclude. This protocol is illustrated in Fig \ref{fig:download}.
}

\subsection{Relaying Manipulated CLU}
We present an extension of FedMobile where clients upload manipulated CLUs via relaying.
\textcolor{black}{Two kinds of manipulation operations are considered. The first is quantization and compression, which aim to reduce the size of the data being transmitted. The second type involves perturbation, implemented to enhance privacy protection.} To avoid confusion, we let $n^t_i$ denote the cumulative gradient update of client $i$'s own, which is not combined with any CLUs received from other clients. We call $n^t_i$ the private-CLU of client $i$. 

\textcolor{black}{
\textbf{Quantization:} The quantization process is denoted by \(Q(\cdot)\), signifying the quantizer operator. To ensure that each local update undergoes quantization exactly once, a client \(i\), upon encountering a suitable relay client \(j\), will quantize its private-CLU \(n^t_i\) instead of the already stored CLU \(m^t_i\). This approach is taken because \(m^t_i\) may include CLUs from others who have used client \(i\) as a relay, and these have been previously compressed. Consequently, during the upload relay from client \(i\) to client \(j\), there is a quantization-related discrepancy represented as \(Q(n^t_i) - n^t_i\), in contrast to scenarios without quantization.
}

\textcolor{black}{
\textbf{Perturbation: } When client \(i\) uploads CLUs to a relay client \(j\), it may incorporate a noise term, such as Gaussian Noise \({\mu}_i^t\), to safeguard its privacy.
}

\textcolor{black}{Although quantization and perturbation effects differ, they can both be conceptualized as introducing a variation term in the original upload transmission. For simplicity and clarity, this difference is denoted as \(\epsilon^t_i\). During the upload relay, if quantization is employed to reduce data size, \(\epsilon^t_i = Q(n^t_i) - n^t_i\) represents the discrepancy introduced by quantization. In contrast, when perturbation is used to increase privacy, \(\epsilon^t_i = {\mu}_i^t\) signifies the added noise.}

\textcolor{black}{\textbf{Upload Timing and Relay Selection.}} This is the same as the original FedMobile strategy. 

\textcolor{black}{\textbf{Upload Relay Mechanism.}} Upon a CLU exchange event at time $t$ between a sender client $i$ and a relay client $j$:

\underline{RESET} (by sender): Before sending the CLU to the relay $j$, the sender first records the noise $\epsilon^t_i$. Then the manipulated CLU, i.e., $\tilde{m}^t_i = m^t_i + \epsilon^t_i$, is sent to the relay. The sender then resets its CLU to $m^t_i:= -\epsilon^t_i$ and its private CLU to $n^t_i:=0$.

\underline{COMBINE} (by relay): 
\textcolor{black}{After receiving $\tilde{m}^t_i$ from client $i$, client $j$ updates it stored CLU $m^t_j$ by incorporating $\tilde{m}^t_i$, i.e., $m^t_j := m^t_j + \tilde{m}^t_i$. }


\textcolor{black}{
\textbf{Remark}:
With the help of relay, the server could get the client $i$'s manipulated CLU $\tilde{m}^t_i = m^t_i + \epsilon^t_i$ sooner. Note that in the case of the manipulated CLU, the sender's CLU is reset to $m^t_i:= -\epsilon^t_i$ instead of $m^t_i:= 0$. This reset ensures that when client $i$ itself reaches the server, its uploaded CLU includes $-\epsilon^t_i$, effectively correcting the previously received manipulated CLU $\tilde{m}^t_i$ back to $m^t_i$. 
}

\section{Convergence Analysis}
\label{section_conv}
\subsection{FedMobile with CLU}
Our analysis utilizes the following assumptions.
\begin{assumption}[Lipschitz Smoothness]\label{assm:smooth}
There exists a constant $L>0$ such that $\|\nabla f_i(x) - \nabla f_i(y)\| \leq L\|x- y\|$, $\forall x, y \in \mathbb{R}^d$ and $\forall i = 1, ..., N$. 
\end{assumption}

\begin{assumption}[Unbiased Local Gradient Estimate]
The local gradient estimate is unbiased, i.e., $\mathbb{E}_\zeta F_i(x, \zeta) = \nabla f_i(x)$, $\forall x$ and $\forall i = 1, \cdots, N$. 
\end{assumption}

\begin{assumption}[Bounded Variance]\label{assm:variance}
There exists a constant $\sigma > 0$ such that $\mathbb{E}[\|\nabla F_i(x, \zeta_i) - \nabla f_i(x)\|^2] \leq \sigma^2$, $\forall x \in \mathbb{R}^d$ and $\forall i = 1, ..., N$. 
\end{assumption}

\begin{assumption}[Bounded Second Moment]\label{assm:sec_moment}
There exists a constant $G > 0$ such that $\mathbb{E}[\|\nabla F_i(x, \zeta_i)\|^2] \leq G^2$, $\forall x \in \mathbb{R}^d$ and $\forall i = 1, ..., N$. 
\end{assumption}

The \textit{real sequence} of the global model is calculated as
\begin{align}
    x^t = x^0 - \frac{1}{N}\sum_{i=1}^N \sum_{s=0}^{\phi_i(t)} \eta g^s_i, ~~~~\forall t
\end{align}
where we define $\phi_i(t)$ to be the time slot up to when all corresponding gradients of client $i$ have been received at time $t$. In the vanilla asynchronous FL case, $\phi_i(t)$  is simply $\tau^\text{last}_i(t)-1$. In FedMobile, $\phi_i(t) > \tau^\text{last}_i(t)-1$ because more information can be uploaded earlier than $t$ due to relaying. 

We also define the \textit{virtual sequence} of the global model, which is achieved in the imaginary ideal case where all local gradients are uploaded to the server instantly at every slot, 
\begin{align}
    v^t = x^0 - \frac{1}{N}\sum_{i=1}^N \sum_{s = 0}^{t-1} \eta g^s_i, ~~~~\forall t
\end{align}

First, we bound the difference $(t-1) - \phi_i(t)$, which characterizes how much CLU information of client $i$ is missing compared to the virtual sequence. 
\begin{lemma}\label{lem:difference}
Assuming at least one semi-qualified upload relay client exists in every upload search interval, then we have $(t - 1) - \phi_i(t) \leq \max\{\Delta - \theta, \Theta\} \triangleq C(\theta, \Theta; \Delta), \forall i = 1,..., N, \forall t$.
\end{lemma}
\begin{proof}
It is obvious that if the server meeting time interval $\tau^\text{next}_i(t) -\tau^\text{last}_i(t) \leq \Theta$, then $(t - 1) - \phi_i(t) = t - \tau^\text{last}_i(t) \leq \Theta$ already holds. Otherwise, for all $t \leq \tau^\text{last}_i(t) + \Theta$, then $(t - 1) - \phi_i(t) = t - \tau^\text{last}_i(t) \leq \Theta$ also holds. Thus, we only need to consider the case $\tau^\text{next}_i(t) -\tau^\text{last}_i(t) > \Theta$ and for time slot $t  > \tau^\text{last}_i(t) + \Theta$. In this case, a semi-qualified relay client is also a qualified relay client because
\begin{align}
    \tau^\text{last}_i(t) + \Theta < \tau^\text{next}_i(t)
\end{align}
By the assumption that at least one semi-qualified relay exists in the search interval, at least one qualified relay must exist. This further implies that the qualified relay client is able to upload a CLU before $t$. Because this CLU contains gradients of client $i$ for at least $\theta$ steps since $\tau^\text{last}_i(t)$, we have $\phi_i(t) \geq \tau^\text{last}_i(t) + \theta - 1$. Therefore, 
\begin{align}
  (t-1) - \phi_i(t) &= \left(t - \tau^\text{last}_i(t)\right) + \left(\tau^\text{last}_i(t) - \phi_i(t) - 1 \right)\nonumber\\
  &\leq \Delta - \theta
\end{align}
To summarize the above cases, $(t - 1) - \phi_i(t) \leq \max\{\Delta - \theta, \Theta\}$ is established. 
\end{proof}


Next, we bound the difference $t - \psi_i(t)$, which characterizes the version difference between the current global model and client $i$'s copy of the global model. 
\begin{lemma}\label{lem:difference2}
Assuming at least one semi-qualified download relay exists in every download search interval, we have $t - \psi_i(t) \leq \max\{\Delta - \omega, \Omega\} \triangleq D(\omega, \Omega; \Delta), \forall i = 1, ..., N, \forall t$.
\end{lemma}
\begin{proof}
Let $t'$ be the meeting time between receiver client $i$ and relay client $j$. Clearly, $\tau^\text{next}_i - \Omega \leq \tau^\text{last}_j(t) \leq t' \leq \tau^\text{next}_i - \omega$ by the definition of the search interval. 

For $t \leq t'$, client $i$ has not met client $j$ yet, so $\psi_i(t) = \tau^\text{last}_i(t)$. Therefore,
\begin{align}
    t - \psi_i(t) &= t - \tau^\text{last}_i(t) \leq t' - \tau^\text{last}_i(t) \nonumber\\&\leq \tau^\text{next}_i(t) - \omega - \tau^\text{last}_i(t) \leq \Delta - \omega
\end{align}

For $t > t'$, client $i$ has met client $j$, so $\psi_i(t) \geq \tau^\text{last}_j(t)$. Therefore,
\begin{align}
    t - \psi_i(t) \leq t - \tau^\text{last}_j(t) \leq t - (\tau^\text{next}_i(t) - \Omega) \leq \Omega
\end{align}
To sum up, $t - \psi_i(t) \leq \max\{\Delta - \omega, \Omega\}$
\end{proof}
The following lemma then bounds the model differences in the real sequence and the virtual sequence. 
\begin{lemma}\label{lem:model_difference}
The difference of the real global model and the virtual global model is bounded as follows
\begin{align}
    \mathbb{E}\left[\|v^t - x^t\|^2\right] \leq C^2(\theta, \Theta; \Delta)\eta^2 G^2
\end{align}
For each client $i$, the difference of its local model and the virtual global model is bounded as follows
\begin{align}
    \mathbb{E}\left[\|v^t - x^t_i\|^2\right] \leq 3(2{D^2(\omega, \Omega;\Delta)} + C^2(\theta, \Theta; \Delta))\eta^2 G^2
\end{align}
\end{lemma}
\begin{proof}
The proof is shown in the supplementary material.
\end{proof}
Now, we are ready to bound the convergence of the real model sequence achieved in FedMobile. 

\begin{theorem}\label{thm:convergence}
Assuming at least one semi-qualified upload (download) relay exists in every upload (download) search interval, by setting $\eta \leq 1/L$, after $T$ time slots, we have
\begin{align}
    &\frac{1}{T}\sum_{t=0}^{T-1} \mathbb{E}\left[\|\nabla f(x^t)\|^2\right]\leq \frac{4}{\eta T}\left(f(x^0) - f^* \right) + \frac{2L\eta\sigma^2}{N} 
    \nonumber\\&+  {4(3D^2(\omega, \Omega;\Delta)} + 2C^2(\theta, \Theta; \Delta))L^2\eta^2G^2 
\end{align}
\end{theorem}
\begin{proof}
The proof is shown in the supplementary material.
\end{proof}
\textbf{Remark}: The convergence bound established in Theorem \ref{thm:convergence} contains three parts. The first part diminishes as $T$ approaches infinity. Both the second and third terms are constants for a constant $\eta$, with the second term depending on our algorithm parameters $\theta$,  $\Theta$, $\omega$ and $\Omega$. For a given final step $T$, one can set $\eta = \frac{\sqrt{N}}{L\sqrt{T}}$ so that the bound becomes
\begin{align}
    &\frac{4L}{\sqrt{NT}}\left(f(x^0) - f^* \right) +  \frac{2\sigma^2}{\sqrt{NT}} \nonumber\\&+ { \frac{4N}{T}(3D^2(\omega, \Omega;\Delta)} + 2C^2(\theta, \Theta; \Delta))G^2
\end{align}
Furthermore, if $T \geq N^3$, the above bound recovers the same $O(\frac{1}{\sqrt{NT}})$ convergence rate of the classic synchronous FL \cite{yu2019parallel} \footnote{There is a subtle difference because $T$ in the asynchronous setting is the number of time slots while in the synchronous setting $T$ is the number of rounds. However, they differ by at most a factor of $\Delta$.}. Next, we discuss the above convergence result in more detail and investigate how the mobility affects the convergence. To this end, we consider for now a fixed client-client meeting rate $\rho$, and investigate how the convergence result depends on our algorithm parameters. 

\underline{Bound Optimization}. Since the convergence bound can be improved by lowering $C(\theta, \Theta; \Delta)$ and $D(\omega, \Omega; \Delta)$, we first investigate them as functions of the algorithm parameters. 
\begin{proposition}\label{prop:timing}
(1) $C(\theta, \Theta; \Delta)$ is non-decreasing in $\Theta$ and non-increasing in $\theta$. Moreover, $\forall \theta, \Theta$, $C(\theta, \Theta; \Delta) \geq \frac{\Delta}{2}$, and the lower bound is attained by choosing $\theta = \Theta = \frac{\Delta}{2}$. (2) $D(\omega, \Omega; \Delta)$ is non-decreasing in $\Omega$ and non-increasing in $\omega$. Moreover, $\forall \omega, \Omega$, $D(\omega, \Omega;\Delta) \geq \frac{\Delta}{2}$, and the lower bound is attained by choosing $\omega = \Omega = \frac{\Delta}{2}$. 
\end{proposition}
Proposition \ref{prop:timing} implies that one should use shorter search intervals, namely $\Theta - \theta$ and $\Omega - \omega$, to lower $C(\theta, \Theta; \Delta)$ and $D(\omega, \Omega; \Delta)$. However, the best $C$ and $D$ are no smaller than $\frac{\Delta}{2}$, which are achieved by choosing $\theta = \Theta = \frac{\Delta}{2}$ and $\omega = \Omega = \frac{\Delta}{2}$. In other words, a short upload search interval around the time $\tau^\text{last}_i + \frac{\Delta}{2}$ and a short download search interval around the time $\tau^\text{next}_i - \frac{\Delta}{2}$ improve the FL convergence. This suggests that the best timing for uploading is at exactly $\tau^\text{last}_i + \frac{\Delta}{2}$ and the best timing for downloading is at exactly $\tau^\text{next}_i - \frac{\Delta}{2}$, which are neither too early nor too late in both cases.

\underline{Probability of Meeting a Semi-Qualified Relay}. The convergence bound in Theorem \ref{thm:convergence}, however, is obtained under the assumption that a client can meet at least one semi-qualified upload (download) relay client in each upload (download) search interval, which may not always hold. How easily a client can find a semi-qualified relay client depends on the {client-client meeting rate. In the VANET example, this rate depends on the D2D communication range.}

Let $Q_u(\theta, \Theta)$ (and $Q_d(\omega, \Omega)$) be the probability that at least one semi-qualified uploading (download) relay is met in an uploading (download) search interval with length $\Theta - \theta$ (and $\Omega - \omega$). They are characterized as follows
\begin{proposition}\label{lem:meeting_prob}
Assuming sufficiently many clients in the system and that clients meet each other uniformly randomly. Let $P_\text{int}(\cdot)$ be the distribution of server meeting intervals. Then $Q_u(\theta, \Theta) = 1 - \prod_{t=0}^{\Theta - \theta}(1 - \rho q_u(\Theta - \theta - t))$ and $Q_d(\omega, \Omega) = 1 - \prod_{t=0}^{\Omega - \omega}(1 - \rho q_d(t))$, where $q_u(\cdot)$ and $q_d(\cdot)$ are distributions computed based on $P_\text{int}(\cdot)$. 
\end{proposition}

\begin{proof}
The proof is shown in the supplementary material.
\end{proof}

Proposition \ref{lem:meeting_prob} states an intuitive result that one should use a larger search interval to increase the probability of meeting a semi-qualified relay. This, however, is not desirable for lowering the convergence bound according to Proposition \ref{prop:timing}. This is exactly where mobility can help improving the FL convergence: by increasing the {client-client meeting rate} $\rho$, a shorter search interval $\Theta - \theta$ (or $\Omega - \omega$) can be used to achieve the same $Q_u$ (or $Q_d$), but a smaller $C$ (or $D$) is obtained. In fact, by relaxing the constraint that a client can only meet one other client at a time slot, with sufficiently many clients in the system, both $Q_u$ and $Q_d$ can approach 1 even if the search interval is just one slot. 

\subsection{FedMobile with Manipulated CLU}
Then we validate the convergence of FedMobile with manipulated CLU uploading. We first state an additional assumption on the noise term $\epsilon^t_i$. 

\begin{assumption}[Bounded Error]\label{assm:error}
The noise term $\epsilon_i$ is bounded, i.e., $\mathbb{E}[\|\epsilon^t_i \|^2] \leq q \mathbb{E}[\|n^t_i \|^2]$, $\forall i = 1, ..., N$, $\forall t$ for some positive real constant $q$. 
\end{assumption}

The \textit{corrupted real sequence} can be written as
\begin{align}
    \widetilde{x}^t = x^t + \frac{1}{N}\sum_{i \in U_t} \epsilon_i, ~~~~\forall t
\end{align}
where we define $U_t$ as the set of clients for whom only corrupted CLUs have been received by the server via the relay, and $|U_t|$ is the size of $U_t$. The real sequence $x^t$ is the imaginary real sequence where the noise is not added. 

\begin{lemma}\label{lem:error_difference}
The difference of the corrupted real global model and the virtual global model is bounded as follows
\begin{align}
    \mathbb{E}\left[\|v^t - \widetilde{x}^t\|^2\right] \leq 2C^2(\theta, \Theta; \Delta)\eta^2 G^2 + 2q\Theta^2\eta^2G^2
\end{align}

For each client $i$, the difference of its local model and the virtual global model is bounded as follows
\begin{align}
    &\mathbb{E}\left[\|v^t - x^t_i\|^2\right] \leq 6({D^2(\omega, \Omega;\Delta)} + C^2(\theta, \Theta; \Delta) + q\Theta^2)\eta^2 G^2
\end{align}
\end{lemma}
\begin{proof}
The proof is shown in the supplementary material.
\end{proof}

\begin{theorem}\label{thm:inacc_convergence}
With manipulated CLU uploading, assuming at least one semi-qualified upload (download) relay client exists in every upload (download) search interval, by setting $\eta \leq 1/L$, after $T$ time slots, we have
\begin{align}
    &\frac{1}{T}\sum_{t=0}^{T-1} \mathbb{E}\left[\|\nabla f(x^t)\|^2\right]\leq \frac{4}{\eta T}\left(f(x^0) - f^* \right) + \frac{2L\eta\sigma^2}{N} 
    \nonumber\\&+  {4(3D^2(\omega, \Omega;\Delta)} + 4C^2(\theta, \Theta; \Delta) + 4q\Theta^2)L^2\eta^2G^2 
\end{align}
\end{theorem}
\begin{proof}
The proof of Theorem \ref{thm:inacc_convergence} follows the proof of Theorem \ref{thm:convergence} by replacing Lemma \ref{lem:model_difference} with Lemma \ref{lem:error_difference}. 
\end{proof}

\textbf{Remark}: With setting $\eta = \frac{\sqrt{N}}{L\sqrt{T}}$, the convergence bound established in Theorem \ref{thm:inacc_convergence} becomes

\begin{align}
    &\frac{4L}{\sqrt{NT}}\left(f(x^0) - f^* \right) +  \frac{2\sigma^2}{\sqrt{NT}} \nonumber\\&+ { \frac{4N}{T}(3D^2(\omega, \Omega;\Delta)} + 4C^2(\theta, \Theta; \Delta) + 4q\Theta^2)G^2
\end{align}
Furthermore, if $T \geq N^3$, the above bound recovers the same $O(\frac{1}{\sqrt{NT}})$ convergence rate of the classic synchronous FL \cite{yu2019parallel}.

\section{The General Case}
\label{sec:general}
\begin{figure}[h]
	\centering
	\begin{minipage}[t]{0.45\linewidth}
	\includegraphics[width=0.99\linewidth]{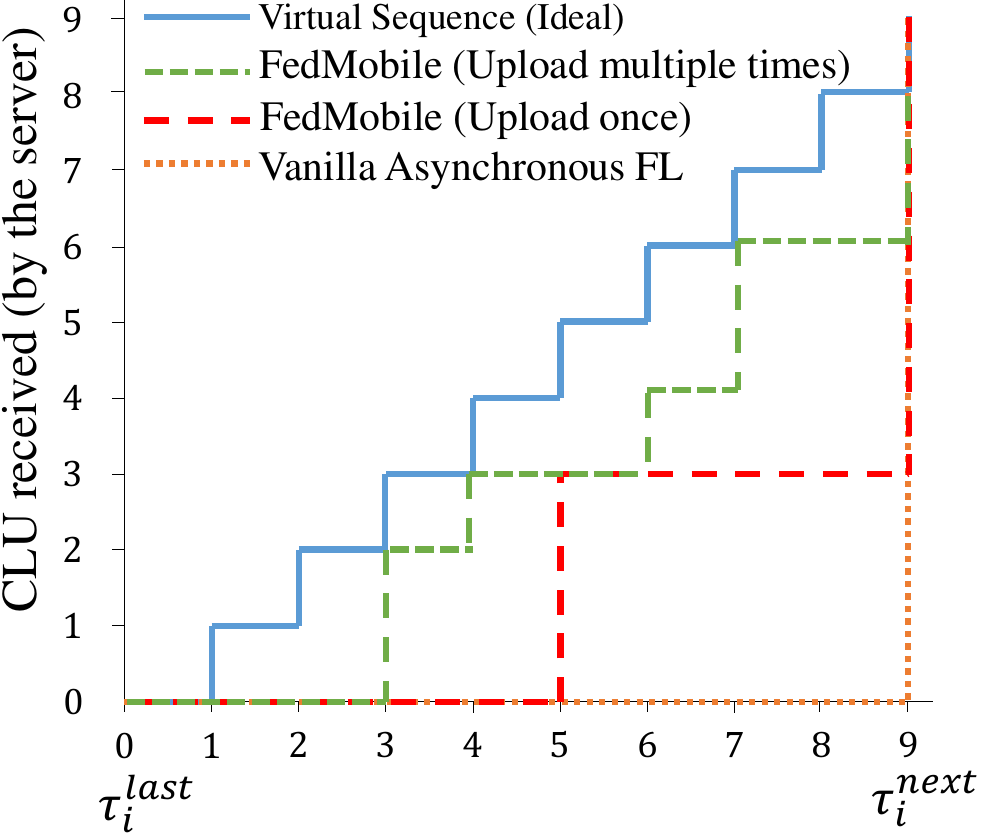}
	\caption{Uploading.} \label{fig:CLUvsTime}
	\end{minipage}
	\begin{minipage}[t]{0.45\linewidth}
	\includegraphics[width=0.99\linewidth]{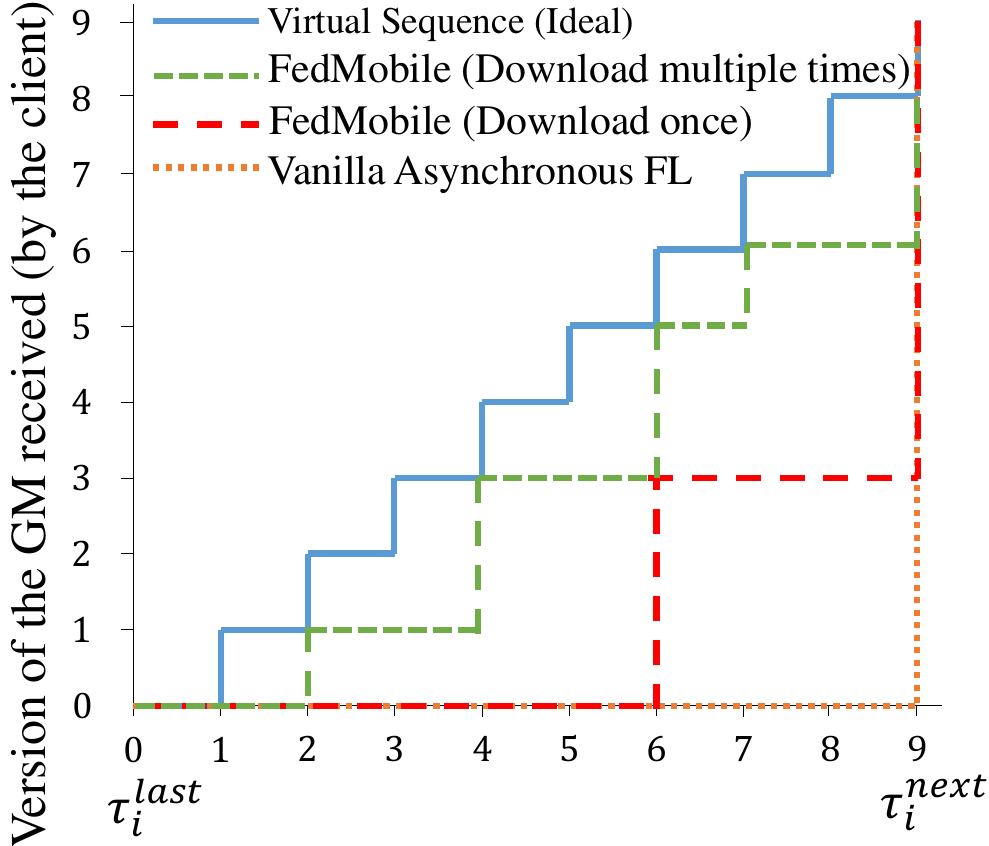}
	\caption{Downloading.} \label{fig:GMvsTime}
	\end{minipage}
\end{figure}


FedMobile can be easily extended to allow multiple upload (download) relay communications between any two consecutive server meetings. We will still have the upload (download) search interval, and FedMobile will simply pick the first $K$ qualified upload (download) relays to perform the upload (download) relay communications. The exact mechanisms of how uploading (downloading) is performed will be slightly changed to handle out-of-order information and avoid redundant information. Our convergence analysis still holds correctly for this generalized case, although the bound may become looser compared to the actual performance. 


Fig. \ref{fig:CLUvsTime}, \ref{fig:GMvsTime} illustrate the key ideas behind FedMobile. Fig. \ref{fig:CLUvsTime} plots hypothetical sequences of CLUs received by the server from a representative client between two consecutive server meetings. The virtual sequence is the ideal case where each local stochastic gradient $g^t_i$ is uploaded to the server instantly after it is computed. Therefore, the received CLU curve is a steady staircase. In the vanilla asynchronous FL, the client uploads the CLU only when it meets the server. Therefore there is a big jump at the next server meeting time but it is all 0 before. FedMobile with at most one upload relaying allows some local stochastic gradient information to be received by the server earlier. The generalized FedMobile allows more CLUs to be relayed and received by the server at earlier time slots. One can imagine that when {$\rho$ is large enough}, it becomes much easier for the client to find qualified relays that can quickly upload CLUs to the server at every time slot. Therefore, the received CLU curve approaches that in the ideal case. Similarly, Fig. \ref{fig:GMvsTime} plots the hypothetical sequence of the global models received by the client.

\section{Experiments}
\begin{figure*}[htbp]
\centering
\subfloat[Performance]{\includegraphics[width=0.24\linewidth]{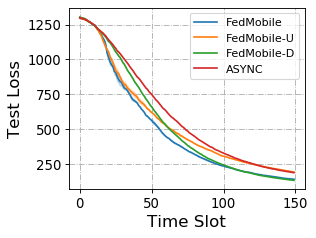}} 
\subfloat[Upload Search Inter.]{\includegraphics[width=0.24\linewidth]{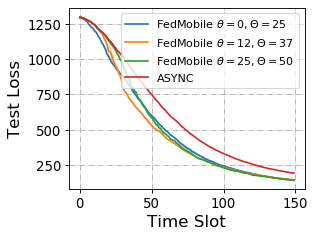}} 
\subfloat[Download Search Inter.]{\includegraphics[width=0.24\linewidth]{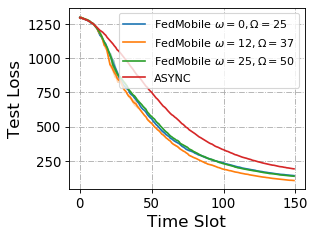}} 
\subfloat[Client Meeting rate]{\includegraphics[width=0.24\linewidth]{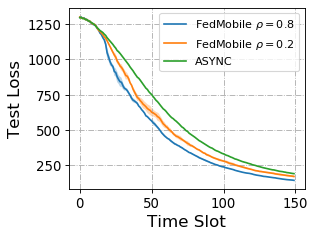}} \\
\subfloat[Multiple Uploads]{\includegraphics[width=0.24\linewidth]{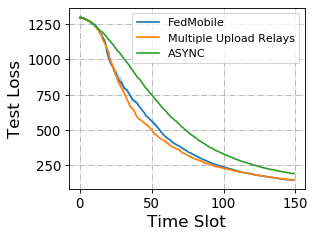}} 
\subfloat[Multiple Downloads]{\includegraphics[width=0.24\linewidth]{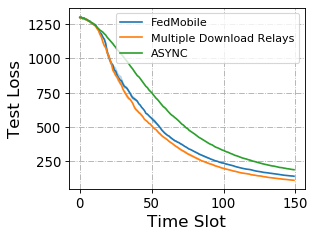}}
\subfloat[Virtual Upload]{\includegraphics[width=0.24\linewidth]{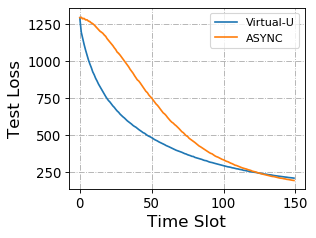}} 
\subfloat[Virtual Downloads]{\includegraphics[width=0.24\linewidth]{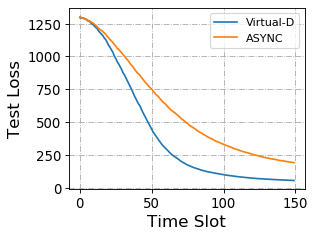}}
\caption{Results on the synthetic data.} \label{synthetic}
\end{figure*}
\subsection{Setup}
We implement FL simulation on the Pytorch framework and perform the model training on one Geforce RTX 3080 GPU. All experiment results are averaged over 3 repeats. To simulate communications among the clients, at each time slot, we uniformly sample $\rho N$ clients over total $N$ clients and randomly construct $\frac{\rho N}{2}$ client pairs. Any two clients in the same pair are simulated to communicate. We conduct experiments on a synthetic dataset and two real-world datasets, i.e., FMNIST \cite{xiao2017fashion} and CIFAR10 \cite{krizhevsky2009learning}. 

 \textbf{Synthetic dataset}: The synthetic dataset is generated on a least-squares linear regression problem. Each data sample has a 200-dimensional feature vector and its real-valued target is calculated as the vector product of the feature and an underlying linear weight plus a 0-mean Gaussian noise. The FL system has 50 clients with each client having 40 data samples. The default training hyper-parameters are: learning rate equals 0.01, learning rate decay factor equals 0.99 until learning rate reaches 0.0001, training batch size equals 128, the total number of training time slots equals 150, default upload parameters $\theta = 10$, $\Theta = 40$ and default download parameters $\omega = 5$, $\Omega = 25$.

\textbf{FMNIST}:The FL system has 50 clients with each client having 400 data samples. We utilize the Dirichlet function ($\alpha = 0.3$) which is typically utilized to simulate the level of non-IID in FL. We use LeNet \cite{lecun1998gradient} as the backbone model. The default hyper-parameters are: learning rate equal to 0.1, learning rate decay factor equals to 0.99 until learning rate reaches 0.001, training batch size equals 128, the total training time slots equals 250, default upload parameters $\theta = 10$, $\Theta = 40$ and default download parameters $\omega = 5$, $\Omega = 25$. 

\textbf{CIFAR10}: The FL system has 50 clients with each client having 600 data samples. The data allocation method is the same as that used for FMNIST. We use ResNet-9 \cite{he2016deep} as the backbone model. The training configuration details are as follows: learning rate equals 0.01, training batch size equals 128, the total number of training time slots equals 500, default upload parameters $\theta = 10$, $\Theta = 40$ and default download parameters $\omega = 5$, $\Omega = 25$. 

\textbf{Benchmarks}. The following benchmarks are considered in our experiments. (1) \textbf{ASYNC}: This is the state-of-the-art asynchronous FL method proposed in \cite{avdiukhin2021federated} to handle arbitrary communication patterns. (2) \textbf{Virtual-U}: This method assumes that each client can upload its local updates immediately to the server at every slot via an imaginary channel but can only download the global model at their actual communication slots. (3) \textbf{Virtual-D}: This method assumes that each client can download the global model from the server at every slot via an imaginary channel but can only upload their CLUs at their actual communication slots. We also consider several variations of FedMobile. (1) \textbf{FedMobile}: This is the proposed algorithm combining both upload and download relay communications. (2) \textbf{FedMobile-U}: This is the proposed algorithm with only upload relaying. (3) \textbf{FedMobile-D}: This is the proposed algorithm with only download relaying. 

\textbf{Communication Patterns}. We consider four distinct communication patterns in total. Three of these patterns simulate asynchronous communication with the server. The \textbf{Fixed-Interval} pattern involves each client $i \in \{1, ..., 50\}$ communicating with the server at intervals defined by the slots $i + 50n, \forall n = 0, 1, 2, ...$. 

For \textbf{Random-Interval}, each client $i$ first communicates with the server at slot $i$, and then continues to communicate at random intervals ranging between 30 and 50 slots.

The \textbf{Random-Interval (Exponential)} pattern is similar, with the first communication of each client $i \in \{1, ..., 50\}$ occurring at slot $i$. However, subsequent communications with the server follow an exponentially distributed pattern with a mean of 30-time slots. The distribution is truncated (with a maximum of 80-time slots) to ensure that the intervals between consecutive server meetings remain bounded. 

Lastly, we simulate a \textbf{Smart Public Transportation} communication pattern, aiming to mimic a real-world public transportation system. This pattern reflects communication structures in four actual bus routes used in Miami, where we envision RSUs being installed.

Our main objective is to enhance the convergence speed in the asynchronous FL setting. It is important to mention that in the experimental results presented below, both FedMobile and the baselines eventually achieve convergence. However, to better highlight the difference in convergence speed between FedMobile and the baselines, we have chosen not to plot the final convergence stage due to the space limitation. We intend to demonstrate that when targeting a specific test accuracy, the utilization of FedMobile leads to a reduction in the required number of time slots.

\subsection{Results on Synthetic Data}
Fig. \ref{synthetic} reports the results (averaged over three repeats) on the synthetic data with a fixed-interval communication pattern. Fig. \ref{synthetic}(a) compares \textbf{FedMobile} and its variations with \textbf{ASYNC} in terms of the test loss. As can been seen, incorporating either upload or download relaying into asynchronous FL improves the FL performance, and a further improvement can be achieved when uploading and downloading are combined. Fig. \ref{synthetic}(b)/(c) illustrates the impact of upload/download search interval on the FL convergence. In both cases, the best search timing is around the middle point of the two consecutive server meeting times, confirming our theoretical analysis in Theorem \ref{thm:convergence} and Proposition \ref{prop:timing}. Fig. \ref{synthetic}(d) shows the impact of $\rho$ on the FL convergence. As predicted by our analysis, a higher $\rho$ in the system improves FL convergence since more timely relay communication opportunities are created. In Figs. \ref{synthetic}(e)(f), we allow clients to use multiple relays to create more communication opportunities with the server whenever possible. The results show that further improvement can indeed be achieved. We also conduct experiments on the ideal relaying scenarios, namely \textbf{Virtual-U} and \textbf{Virtual-D}, to illustrate what can be achieved in the ideal case. The results verify our hypothesis that more communication opportunities with the server benefit convergence. It is also interesting to note that virtual uploading/downloading has a more significant impact on the early/late FL slots, suggesting that an adaptive design may better balance the FL performance and the resource cost. 


\subsection{Results on CIFAR10}
\begin{figure}[h]
\vspace{-10pt}
\centering
\subfloat[CIFAR10 (Fixed)]{\includegraphics[width=0.49\linewidth]{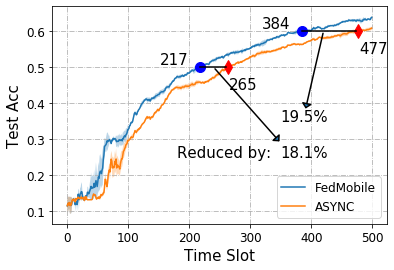}} 
\subfloat[CIFAR10 (Random)]{\includegraphics[width=0.49\linewidth]{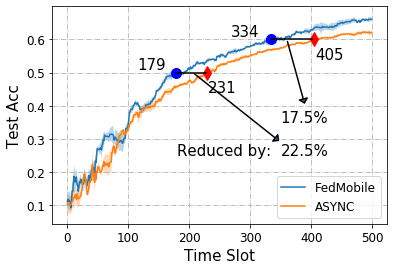}} 
\caption{Performance Comparison on CIFAR10.} \label{cifar-performance}
\end{figure}

We now report the results (averaged over three repeats) on CIFAR10. We validate that our proposed method, \textbf{FedMobile} can achieve a better convergence speed than \textbf{ASYNC} on CIFAR10 under both Fixed-Interval and Random-Interval settings in Fig. \ref{cifar-performance}. Note that the main point of our paper is utilizing relays to improve the convergence speed. Hence, by setting a target accuracy level and comparing the required number of time slots to reach it, the notable advancements of FedMobile are evident. For a specific example, to achieve 60\% accuracy under fixed-interval setting, \textbf{ASYNC} needs about 384 slots while \textbf{FedMobile} only needs 477 slots. The required time slot decreases by 19.5\%. 

\vspace{-10pt}
\subsection{Results on FMNIST}
\begin{figure}[h]
\centering
\subfloat[\textcolor{black}{FMNIST (Fixed)}]{\includegraphics[width=0.49\linewidth]{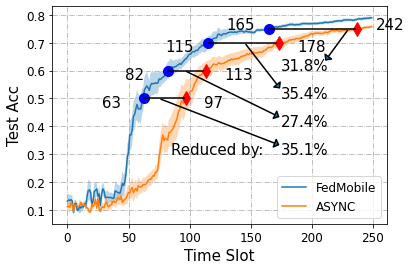}} 
\subfloat[\textcolor{black}{FMNIST (Random)}]{\includegraphics[width=0.49\linewidth]{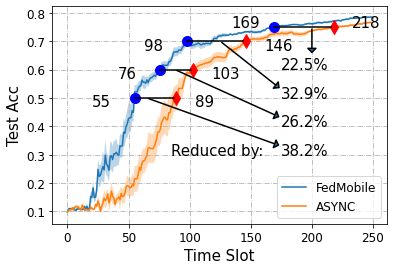}} 
\caption{\textcolor{black}{Performance Comparison on FMNIST.}} 
\label{minist-performance}
\end{figure}

Then we report the results (averaged over three repeats) on FMNIST. Fig.~\ref{minist-performance} shows that \textbf{FedMobile} achieves a better convergence speed than \textbf{ASYNC} on \textcolor{black}{FMNIST} under both the Fixed-Interval and Random-Interval settings. 

\begin{figure}[h]
\centering
\vspace{-10pt}
\subfloat[\textcolor{black}{Larger communication intervals}]{\includegraphics[width=0.48\linewidth]{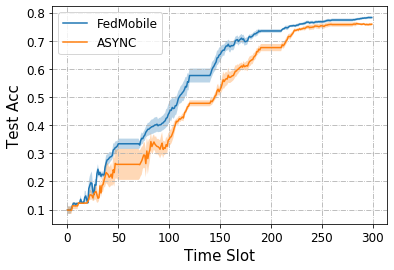}}
\subfloat[\textcolor{black}{Larger amount of clients}]{\includegraphics[width=0.48\linewidth]{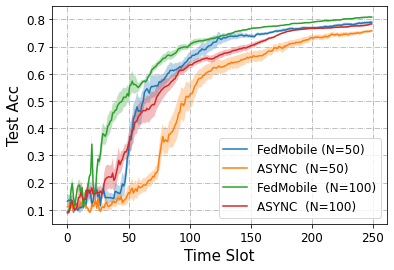}} 
\caption{\textcolor{black}{Performance Comparisons Under Various Configs.}} \label{fig: sa}
\end{figure}

\textcolor{black}{
\textbf{Scalability Analysis}: 
To validate the scalability of our proposed FedMobile method, we conduct supplementary experiments under a Fixed-Interval setting. Initially, we set the interval at 50-time slots, aiming to evaluate whether FedMobile maintains its superiority over the baseline method as the interval increases. With a constant number of clients, we extend the interval to 70-time slots and adjust the upload/download search intervals accordingly (i.e., $\theta = 20, \Theta = 60, \omega = 10, \Omega = 30$). The outcomes, depicted in Figure \ref{fig: sa} (a), affirm that FedMobile surpasses the baseline method.
}

\textcolor{black}{
Furthermore, we validate FedMobile's performance with an increased client count. Maintaining the fixed interval at 50-time slots, we augment the client number from 50 to 100, with each client possessing 400 data samples exhibiting a non-IID degree of $\alpha = 0.3$. The findings, showcased in Figure \ref{fig: sa} (b), reveal that FedMobile consistently exceeds the baseline method. Notably, the inclusion of more clients accelerates the training process for both FedMobile and the baseline method, with FedMobile continually demonstrating superior performance. These experiments collectively attest to the scalability of FedMobile across diverse configurations.}



\begin{figure}[h]
\centering
\vspace{-10pt}
\subfloat[Upload/Download (Fixed)]{\includegraphics[width=0.48\linewidth]{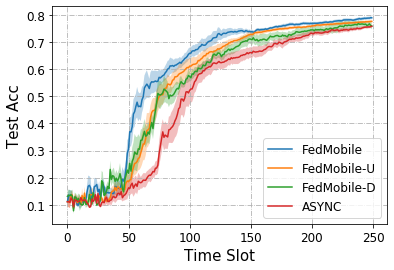}} 
\subfloat[Upload/Download (Random)]{\includegraphics[width=0.48\linewidth]{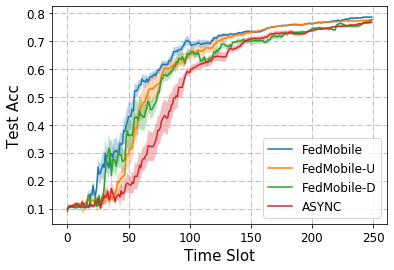}} 
\caption{Effect of Upload/Download Relay.} \label{Upload-Download}
\end{figure}

\textbf{Effect of Upload/Download}: \textcolor{black}{Under both communication patterns,} Fig. \ref{Upload-Download}(a) and (b) demonstrate that while upload and download relay communications individually can improve the FL convergence performance, combining them results in an additional benefit.

\begin{figure}[h]
\centering
\subfloat[Multiple Upload (Fixed)]{\includegraphics[width=0.49\linewidth]{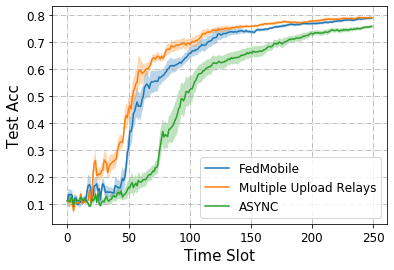}} 
\subfloat[Multiple Upload (Random)]{\includegraphics[width=0.49\linewidth]{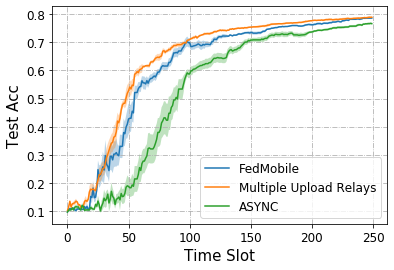}} 
\caption{Effect of Multiple Upload Relays.} \label{Mult-Upload}
    \vspace{-10 pt}
\end{figure}

\textbf{Multiple Relays}: In the previous result of the synthetic dataset, we find that both multiple uploads relays and multiple downloads relays can improve FedMobile's convergence speed. In the real-world dataset experiments, Fig. \ref{Mult-Upload}(a) and (b) show that using multiple upload relay communications further improves the FL convergence performance. However, we observed in our experiments that this is not the case with using multiple download relay communications. This is likely due to the complexity of real-world datasets, which causes the assumptions for our theoretical analysis to be violated. This represents a limitation of our current analysis and requires further investigation.

\begin{figure}[h]
\vspace{-10pt}
\centering
\subfloat[Virtual-U (Fixed)]{\includegraphics[width=0.49\linewidth]{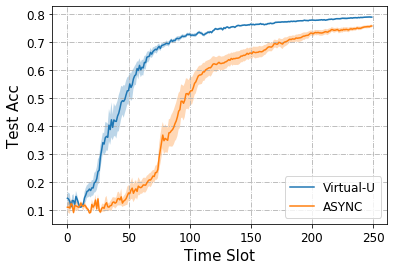}} 
\subfloat[Virtual-U (Random)]{\includegraphics[width=0.49\linewidth]{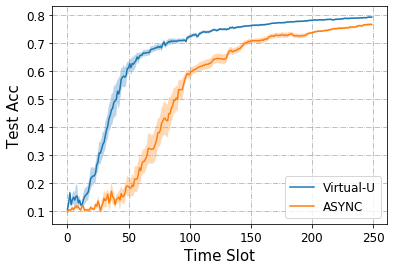}} 
\caption{Virtual-U on FMNIST.} \label{virtual-U}
\end{figure}

\begin{figure}[h]
\centering
\vspace{-20pt}
\subfloat[Virtual-D (Fixed)]{\includegraphics[width=0.49\linewidth]{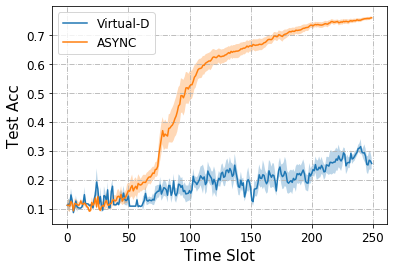}} 
\subfloat[Virtual-D (Random)]{\includegraphics[width=0.49\linewidth]{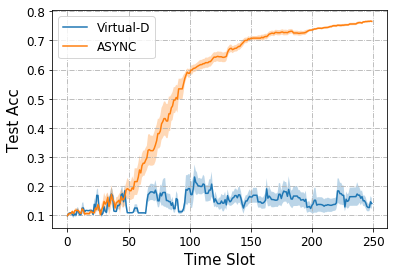}} 
\caption{Virtual-D on FMNIST.} \label{virtual-D}
\end{figure}

\textbf{Virtual-U and Virtual-D}: \textcolor{black}{To further present the different effects between multiple uploads and multiple downloads, we consider two ideal cases.} \textbf{Virtual-U} and \textbf{Virtual-D} work as the ideal cases for upload relaying and download relaying and hence we conduct experiments to investigate their performance on the real-world dataset. Figs. \ref{virtual-U} and \ref{virtual-D} report the performance of \textbf{Virtual-U} and \textbf{Virtual-D} on FMNIST with both fixed and random interval communication patterns. Similar to Fig. \ref{synthetic}(g) for the synthetic data, Figs. \ref{virtual-U} (a)(b) show that \textbf{Virtual-U} can greatly improve the convergence performance. 
However, different from Fig. \ref{synthetic}(h) for the synthetic data, Figs. \ref{virtual-D} (a)(b) show that \textbf{Virtual-D} fail to converge on the real-world dataset. We conjecture that this is likely due to the complexity of the real-world dataset where some of the assumptions needed for our theoretical analysis do not strictly hold. However, as Fig. \ref{Upload-Download} shows, using one-time download relaying still has benefits on the convergence even in the real-world dataset. 

\begin{figure}[h]
\vspace{-10pt}
\centering
\subfloat[Upload Search Inter.]{\includegraphics[width=0.49\linewidth]{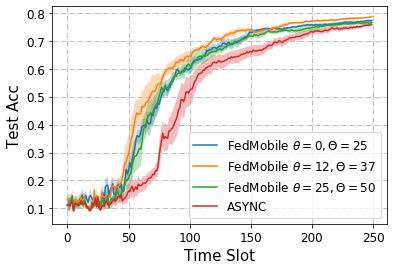}} 
\subfloat[Download Search Inter.]{\includegraphics[width=0.49\linewidth]{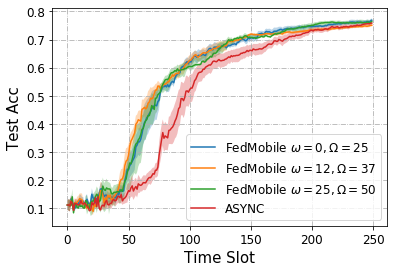}}
\caption{Effect of upload/download Interval} \label{Timing}
\end{figure}

\textbf{Setting of upload/download interval}: Our theoretical analysis emphasizes the importance of precise timing in both the upload and download relay mechanisms. To achieve optimal performance and avoid relay operations that occur too early or too late, it is crucial to carefully set the parameters for the upload/download search interval. These parameters should be around the median value of the maximum time interval (denoted as $\Delta$) between consecutive server communications. This concept is demonstrated in Fig.~\ref{Timing}, where adjusting the upload/download parameters around the midpoint of $\Delta$ yields relay timings that result in the best convergence speed. 

\begin{figure}[h]
\vspace{-10pt}
    \centering
    \begin{minipage}[t]{0.49\linewidth}
        \includegraphics[width=0.99\linewidth]{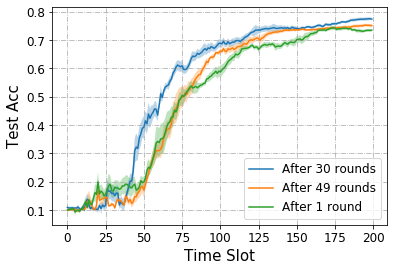}
        \caption{Download Time}
	    \label{download-time}
    \end{minipage}
     \begin{minipage}[t]{0.49\linewidth}
        \includegraphics[width=0.99\linewidth]{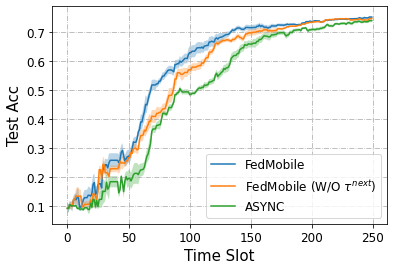}
        \caption{\textcolor{black}{Unknown Next Meeting Time under the \textbf{Random-Interval (Exponential)}}}
	    \label{exponetial}
    \end{minipage}
\vspace{-10pt}
\end{figure}

However, it's worth noting the marginal variability among the three download search intervals depicted in Fig.~\ref{Timing}(b). To better comprehend the impact of download relay timing, we devised a hypothetical scenario wherein each client could download the global model 1/30/49 time slot(s) after their last server interaction. The outcome, as illustrated in Fig.~\ref{download-time}, unambiguously demonstrates that the optimal download relay timing should steer clear of being too early or too late.

\textcolor{black}{
\textbf{Unknown Next meeting time}:  
We further evaluate our proposed method using the \textbf{Random-Interval (Exponential)} pattern. We here
analyze two distinct scenarios: one in which clients are informed of their next meeting time with the server, and another where this time is not known. In the latter case, the expected value of the exponential distribution is used to predict the timing of the next server interaction. As illustrated in Fig. \ref{exponetial}, our findings reveal that FedMobile consistently outperforms the baseline, regardless of whether it employs precise next meeting time data or estimates based on the expected value of the exponential distribution.
}

\begin{figure}[h]
\vspace{-10pt}
\centering
\subfloat[Client Meeting Rate (Fixed)]{\includegraphics[width=0.49\linewidth]{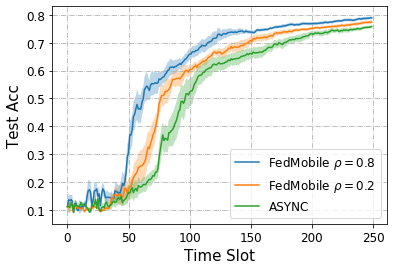}} 
\subfloat[Client Meeting Rate (Random)]{\includegraphics[width=0.49\linewidth]{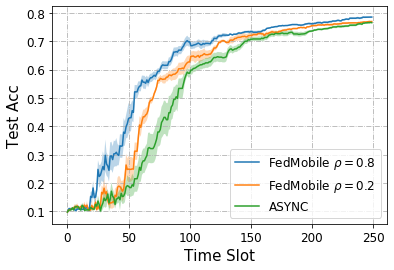}} 
\caption{Client Meeting Rate Effect.} \label{Meeting}
\end{figure}

\textbf{Client-Client Meeting Rate}:
We further investigate the effect of {client-client meeting rate} under the fixed and random-interval communication pattern. Fig. \ref{Meeting}(a)(b) shows that the higher {client-client meeting rate} improves convergence speed.

\begin{figure}[h]
    \centering
    \begin{minipage}[t]{0.49\linewidth}
	\includegraphics[width=1\linewidth]{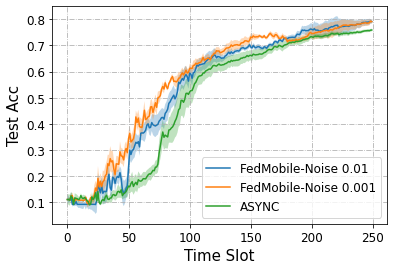}
	\caption{Noise Level} \label{fig:DP}
    \end{minipage}
    \begin{minipage}[t]{0.49\linewidth}
    \includegraphics[width=1\linewidth]{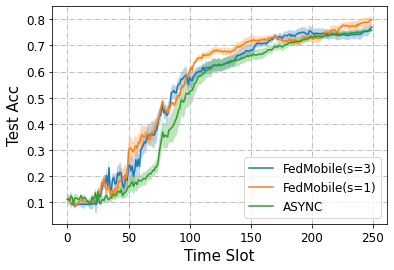}
	\caption{Quantization} \label{fig:QT}   
    \end{minipage} 
\end{figure}

\textbf{Relaying Manipulated CLU:}
We test two types of manipulation. In the first type, we directly add Gaussian noises to the relayed CLU. Fig. \ref{fig:DP} shows the convergence curves under different amount of noises. (e.g.  Gaussian Noise $\mathcal{N}_1(0, 0.01)$ and Gaussian Noise $\mathcal{N}_2(0, 0.001)$). In the second type, we utilize the low precision quantizer in \cite{alistarh2017qsgd} to quantize the CLU before relaying. Here the quantization level is defined as $s$. Fig. \ref{fig:QT} shows the convergence curves under different quantization levels. In both cases, FedMobile still outperforms the baseline method provided that the added noise is small or the adopted quantization level is low. 

\begin{figure}[h]
    \centering
    \begin{minipage}[t]{0.49\linewidth}
	\includegraphics[width=1\linewidth, height =0.7\linewidth]{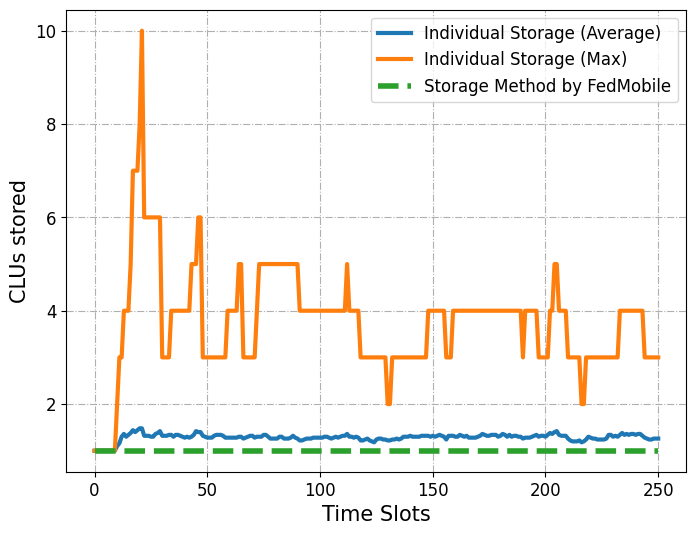}
	\caption{\textcolor{black}{Storage Efficient}} \label{fig:storage}
    \end{minipage}
     \begin{minipage}[t]{0.49\linewidth}
	\includegraphics[width=1\linewidth, height =0.7\linewidth]{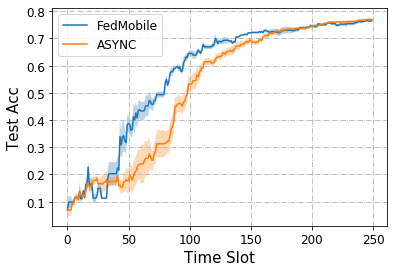}
	\caption{Performances (Smart Publication Transportation)} \label{fig:smart_result}
    \end{minipage}
\end{figure}

\textcolor{black}{
\textbf{Storage Efficient} In FedMobile, we employ the \underline{RESET} and \underline{COMBINE} actions to establish a streamlined and storage-efficient method that ensures precise, one-time delivery of specific information to the server, without retaining client ID information. Compared to the approach of recording each received client ID and storing each local update separately at the relay client, FedMobile's storage mechanism exhibits significantly higher efficiency, as illustrated in Fig \ref{fig:storage}.
}

\begin{figure}[h]
    \centering
    \begin{minipage}[t]{0.9\linewidth}
	\includegraphics[width=0.9\linewidth]{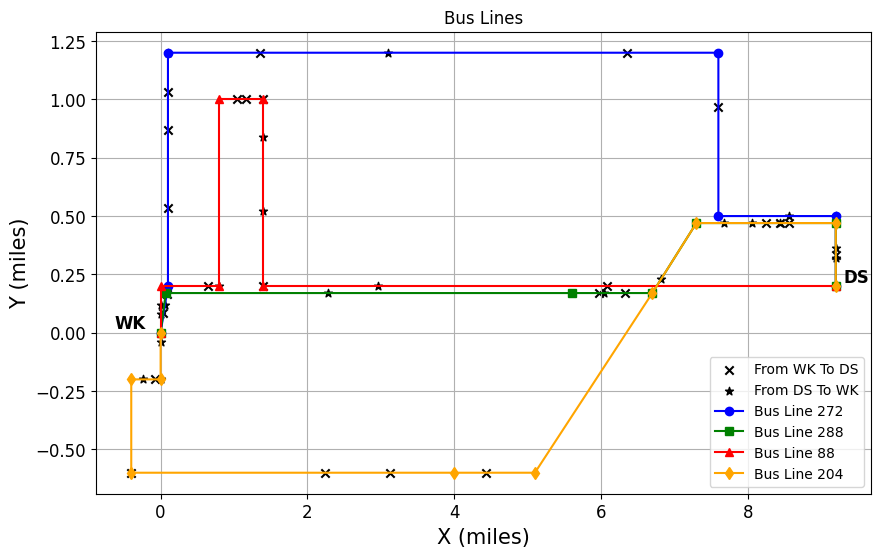}
	\caption{An illustration of smart public transportation is provided, using four real bus lines in Miami.} \label{fig:smart}
    \end{minipage}
\end{figure}

\textbf{FedMobile in Smart Publication Transportation:} We here contemplate implementing our suggested method, FedMobile, in a more real-world communication pattern scenario, such as smart public transportation. This pattern embodies communication structures seen in four actual bus routes operating in Miami between West Kendall (WK) and Dadeland South (DS), assuming the installation of RSUs. For simplicity, and without losing generality, we posit that two RSUs are positioned at the two destinations (West Kendall and Dadeland South). We hypothesize that on each bus line, there are 12/13 buses (totaling 50 buses for all 4 lines) that begin from different starting points at the commencement of the training process. On each line, 5 buses initiate their route from West Kendall to Dadeland South, while the remaining 5 start from Dadeland South heading towards West Kendall. Each bus will switch its direction upon reaching the terminal (either Dadeland South or West Kendall) and will continue operating on its route throughout the entire training process. This arrangement is visually represented in Fig. \ref{fig:smart}. Aligning with the real-world settings, we consider the buses to travel at a speed between 0.4 to 0.6 miles per minute. The time slot, defined as the duration required for one round of local training, is set to 30 seconds. The V2V communication range is fixed at 0.6 miles. To simulate the training, we employ the local client data distribution as described by FMNIST in the setup section. The outcomes presented in Fig. \ref{fig:smart_result} confirm that our recommended method, FedMobile, surpasses the performance of the state-of-the-art Asynchronous FL method, ASYNC.

\section{Conclusion}
This paper focuses on the design of asynchronous Federated Learning (FL) algorithms for practical systems where continuous client-server communication is not always available. We emphasize the importance of client mobility and the resulting random client-client communication opportunities in facilitating timely information exchange for model training in asynchronous FL. To harness the advantages of additional client-client communication, we propose a new FL algorithm called FedMobile. FedMobile not only accelerates convergence but also maintains storage efficiency and prevents duplicate update transmission. We provide a detailed convergence analysis and conduct extensive experiments to validate our proposed method. The results demonstrate that FedMobile significantly improves convergence speed. We believe that FedMobile has the potential to advance distributed machine learning, especially FL, in various real-world systems such as mobile gaming, mobile sensing, and smart vehicular networks.

\bibliographystyle{IEEEtran}
\bibliography{reference}

\clearpage
\newpage
\appendices
\section{Proof of Lemma \ref{lem:model_difference}}
\label{proof-l3}
Consider any time $t$, by the definition of virtual and real sequences, we have
\begin{align}
    &\mathbb{E}\left[\|v^t - x^t\|^2\right] = \mathbb{E}\left[\left\|\frac{1}{N}\sum_{i=1}^N \sum_{s= \phi_i(t)}^{t-1} \eta g^s_i\right\|^2\right]\nonumber\\
    \leq&\frac{\eta^2}{N}\sum_{i=1}^N \mathbb{E}\left[\left\|\sum_{s= \phi_i(t)}^{t-1} g^s_i\right\|^2\right]  \nonumber\\ \leq& \frac{\eta^2}{N}\sum_{i=1}^N((t-1) - \phi_i(t))^2G^2\nonumber\\
    \leq& C^2(\theta, \Theta; \Delta)\eta^2 G^2
\end{align}
where the first $\leq$ is due to the Cauchy-Schwarz inequality, the second $\leq$ is due to both the Cauchy-Schwarz inequality and Assumption 4, and the last $\leq$ is due to Lemma \ref{lem:difference}.

Consider client $i$ at any time $t$ with the global model version $\psi_i(t)$. Therefore client $i$ is doing the local training steps using the global model $x^{\psi_i(t)}$ as the initial model. Thus, we use $x^{\psi_i(t)}$ as the anchor model to investigate the difference between $v^t$ and $x^t_i$. 
\begin{align}
    &\mathbb{E}\left[\left\|v^t - x^{t}_i\right\|^2\right] \nonumber\\=& \mathbb{E}\left[\left\|v^t - v^{\psi_i(t)} + v^{\psi_i(t)} - x^{\psi_i(t)} + x^{\psi_i(t)} - x^{t}_i\right\|^2\right]\nonumber\\
    \leq& 3\mathbb{E}\left[\left\|v^t - {v^{\psi_i(t)}}\right\|^2 + \left\|v^{\psi_i(t)} - x^{\psi_i(t)}\right\|^2 + \left\|x^{\psi_i(t)} - x^{t}_i\right\|^2\right]\nonumber\\
    \leq& 3\mathbb{E}\left[\left\|\frac{1}{N}\sum_{i=1}^N \sum_{s= \psi_i(t)}^{t-1} \eta g^s_i\right\|^2\right] + 3\mathbb{E}\left[\left\|v^{\psi_i(t)} - x^{\psi_i(t)}\right\|^2\right] \nonumber\\ &+ 3\mathbb{E}\left[\left\|x^{t'}_i - x^t_i\right\|^2\right] \nonumber\\
    \leq& 3\mathbb{E}\left[\left\|\frac{1}{N}\sum_{i=1}^N \sum_{s= \psi_i(t)}^{t-1} \eta g^s_i\right\|^2\right] + 3\mathbb{E}\left[\left\|v^{\psi_i(t)} - x^{\psi_i(t)}\right\|^2\right] \nonumber\\&+ 3\mathbb{E}\left[\left\|x^{\psi_i(t)}_i - x^t_i\right\|^2\right]\nonumber\\
    \leq &3D^2(\omega, \Omega;\Delta)\eta^2 G^2 +3C^2(\theta, \Theta; \Delta)\eta^2 G^2 \nonumber\\&+ 3{D^2(\omega, \Omega;\Delta)}\eta^2G^2\nonumber\\
    =& 3({2D^2(\omega, \Omega;\Delta)} + C^2(\theta, \Theta; \Delta))\eta^2 G^2,
\end{align}
\textcolor{black}{where the first $\leq$ is due to the Cauchy-Schwarz inequality, the second $\leq$ is due to the definition of virtual global model, the third $\leq$ is due to the fact that $t - t' \leq t - \psi_i(t)$ always holds.} If the client $i$ has not met its download relay, $t' = \psi_i(t)$. Otherwise (the client $i$ has met its download relay), $t'$ is the meeting time and therefore, $t' > \psi_i(t)$. Note that this bound holds even if a relay client $j$ simply sends its local model $x^s_j$ to the client $i$ at some time $t \leq \psi(t) \leq s \leq t$. 

\section{Proof of Theorem \ref{thm:convergence}}
\label{proof-t1}
We first analyze the convergence of the virtual global model sequence. By the smoothness of $f(x)$ (Assumption \ref{assm:smooth}), we have
\begin{align}\label{f_bound}
    &\mathbb{E}[\|f(v^{t+1})\|] \nonumber\\\leq& \mathbb{E}[\|f(v^t)\|] + \mathbb{E}[\langle \nabla f(v^t), v^{t+1} - v^t\rangle] + \frac{L}{2} \mathbb{E}[\|v^{t+1} - v^t\|^2]
\end{align}
The second term on the right-hand side of \eqref{f_bound} can be expressed as follows,
\begin{align}
    &\mathbb{E}[\langle \nabla f(v^t), v^{t+1} - v^t\rangle] \nonumber\\
    =& -\eta \mathbb{E}[\langle \nabla f(v^t), \frac{1}{N}\sum_{i=1}^N g^t_i\rangle] \nonumber\\=& -\eta \mathbb{E}[\langle \nabla f(v^t), \frac{1}{N}\sum_{i=1}^N \nabla f_i(x^t_i)\rangle]\nonumber\\
    =&-\frac{\eta}{2}\mathbb{E}\left[\|\nabla f(v^t)\|^2\right] -\frac{\eta}{2}\mathbb{E}\left[ \|\frac{1}{N}\sum_{i=1}^N \nabla f_i(x^t_i)\|^2\right] \nonumber\\&+\frac{\eta}{2}\mathbb{E}\left[ \|\nabla f(v^t) - \frac{1}{N}\sum_{i=1}^N \nabla f_i(x^t_i)\|^2\right]
\end{align}


The third term on the right-hand side of \eqref{f_bound} can be bounded as follows,
\begin{align}
    &\frac{L}{2} \mathbb{E}[\|v^{t+1} - v^t\|^2] = \frac{L\eta^2}{2}\mathbb{E}\left[\|\frac{1}{N}\sum_{i=1}^N g^t_i\|^2\right]\nonumber\\
    = & \frac{L\eta^2}{2} \mathbb{E}\left[\|\frac{1}{N}\sum_{i=1}^N (g^t_i - \nabla f_i(x^t_i))\|^2 \right] \nonumber\\&+ \frac{L\eta^2}{2}\mathbb{E}\left[\|\frac{1}{N}\sum_{i=1}^N  \nabla f_i(x^t_i)\|^2 \right]\nonumber\\
    =&\frac{L\eta^2}{2N^2}\sum_{i=1}^N \mathbb{E}\left[\|g^t_i - \nabla f_i(x^t_i)\|^2\right] \nonumber\\&+ \frac{L\eta^2}{2}\mathbb{E}\left[\|\frac{1}{N}\sum_{i=1}^N  \nabla f_i(x^t_i)\|^2 \right]\nonumber\\
    \leq& \frac{L\eta^2\sigma^2}{2N} + \frac{L\eta^2}{2}\mathbb{E}\left[\|\frac{1}{N}\sum_{i=1}^N  \nabla f_i(x^t_i)\|^2 \right]
\end{align}
Substituting these into \eqref{f_bound} yields
\begin{align}
    &\mathbb{E}[\|f(v^{t+1}\|] \nonumber\\
    \leq& \mathbb{E}[\|f(v^t)\|]-\frac{\eta}{2}\mathbb{E}\left[\|\nabla f(v^t)\|^2\right] \nonumber\\&-\frac{\eta - L\eta^2}{2}\mathbb{E}\left[ \|\frac{1}{N}\sum_{i=1}^N \nabla f_i(x^t_i)\|^2\right] \nonumber\\
    &+\frac{\eta}{2}\mathbb{E}\left[ \|\nabla f(v^t) - \frac{1}{N}\sum_{i=1}^N \nabla f_i(x^t_i)\|^2\right] +\frac{L\eta^2\sigma^2}{2N}\nonumber\\
    \leq& \mathbb{E}[\|f(v^t)\|]-\frac{\eta}{2}\mathbb{E}\left[\|\nabla f(v^t)\|^2\right]\nonumber\\&+\frac{\eta}{2}\mathbb{E}\left[ \|\nabla f(v^t) - \frac{1}{N}\sum_{i=1}^N \nabla f_i(x^t_i)\|^2\right] +\frac{L\eta^2\sigma^2}{2N}\nonumber\\
    \leq& \mathbb{E}[\|f(v^t)\|]-\frac{\eta}{2}\mathbb{E}\left[\|\nabla f(v^t)\|^2\right]\nonumber\\&+\frac{\eta}{2N}\sum_{i=1}^N\mathbb{E}\left[\|\nabla f(v^t) - \nabla f_i(x^t_i)\|^2\right] +\frac{L\eta^2\sigma^2}{2N}\nonumber\\
    \leq& \mathbb{E}[\|f(v^t)\|]-\frac{\eta}{2}\mathbb{E}\left[\|\nabla f(v^t)\|^2\right]\nonumber\\&+\frac{\eta L^2}{2N}\sum_{i=1}^N\mathbb{E}\left[\|v^t - x^t_i\|^2\right] +\frac{L\eta^2\sigma^2}{2N} \nonumber\\
    \leq& \mathbb{E}[\|f(v^t)\|]-\frac{\eta}{2}\mathbb{E}\left[\|\nabla f(v^t)\|^2\right] \nonumber\\&+ ({3D^2(\omega, \Omega;\Delta)} + 1.5C^2(\theta, \Theta; \Delta))L^2\eta^3G^2 + \frac{L\eta^2\sigma^2}{2N}
\end{align}
Dividing both sides by $\frac{\eta}{2}$ and rearranging the terms, we have
\begin{align}
    \mathbb{E}\left[\|\nabla f(v^t)\|^2\right] &\leq \frac{2}{\eta}\left(\mathbb{E}[\|f(v^t)\|] - \mathbb{E}[\|f(v^{t+1}\|] \right)  \nonumber\\
    &+ 3({2D^2(\omega, \Omega;\Delta)} + C^2(\theta, \Theta; \Delta))L^2\eta^2G^2 \nonumber\\&+ \frac{L\eta\sigma^2}{N}
\end{align}
Taking the sum over $t = 0, 1, ..., T-1$ and dividing both sides by $T$, we have
\begin{align}
    &\frac{1}{T}\sum_{t=0}^{T-1} \mathbb{E}\left[\|\nabla f(v^t)\|^2\right] \nonumber\\
    \leq & \frac{2}{\eta T}\left(\mathbb{E}[\|f(v^0)\|] - \mathbb{E}[\|f(v^T\|] \right) \nonumber\\&+ 3({2D^2(\omega, \Omega;\Delta)} + C^2(\theta, \Theta; \Delta))L^2\eta^2G^2 + \frac{L\eta\sigma^2}{N}\nonumber\\
    \leq & \frac{2}{\eta T}\left(f(x^0) - f^* \right) \nonumber\\&+ 3({2D^2(\omega, \Omega;\Delta)} + C^2(\theta, \Theta; \Delta))L^2\eta^2G^2 + \frac{L\eta\sigma^2}{N}
\end{align}
where $f^* = \min_x f(x)$. 
Now, for the real sequence, we have
\begin{align}
    &\frac{1}{T}\sum_{t=0}^{T-1} \mathbb{E}\left[\|\nabla f(x^t)\|^2\right]\nonumber\\
    \leq & \frac{1}{T}\sum_{t=0}^{T-1} (2\mathbb{E}\left[\|\nabla f(v^t)\|^2\right] + 2\mathbb{E}\left[\|\nabla f(v^t) - \nabla f(x^t)\|^2\right])\nonumber\\
    \leq & \frac{1}{T}\sum_{t=0}^{T-1} (2\mathbb{E}\left[\|\nabla f(v^t)\|^2\right] + 2L^2\mathbb{E}\left[\|v^t - x^t\|^2\right])\nonumber\\
    \leq & \frac{2}{T}\sum_{t=0}^{T-1} \mathbb{E}\left[\|\nabla f(v^t)\|^2\right] + 2C^2(\theta, \Theta; \Delta)L^2\eta^2G^2
\end{align}
Substituting the bound on the virtual sequence into the above equation, we have,
\begin{align}
    &\frac{1}{T}\sum_{t=0}^{T-1} \mathbb{E}\left[\|\nabla f(x^t)\|^2\right] \nonumber\\\leq& \frac{4}{\eta T}\left(f(x^0) - f^* \right) + 4({3D^2(\omega, \Omega;\Delta)}  \nonumber\\& + 2C^2(\theta, \Theta; \Delta))L^2\eta^2G^2 + \frac{2L\eta\sigma^2}{N}
\end{align}
This completes the proof.

\section{Proof of Proposition \ref{lem:meeting_prob}}
\label{proof-p2}
Let $\pi_u(s)$ be the fraction of clients whose next server meeting time is in $s$ time slots. For sufficiently many clients, the distribution $\pi_u = [\pi_u(0)~\pi_u(1)~...~\pi_u(\Delta)]$ is time-invariant and can be calculated by solving the stationary distribution of a Markov chain that describes the client state in terms of the remaining server meeting time. Specifically, $\pi_u$ is the solution to
\begin{align}
    \pi_u
    \begin{bmatrix}
    0 & P_\text{int}(1) & P_\text{int}(2) & ... & P_\text{int}(\Delta)\\
    1 & 0 & 0 & \cdots & 0\\
    0 & 1 & 0 & \cdots & 0\\
    \vdots & \ddots & \ddots & \ddots & \vdots\\
    0 & \cdots & 0 & 1 & 0
    \end{bmatrix}
    = \pi_u
\end{align}
Furthermore, we let $q_u(t) = \sum_{s = 0}^t \pi_u(s)$ be the cumulative distribution function. Suppose in the $(t+1)$-th slot in the upload search interval, a client is met. The probability that this client is a semi-qualified upload relay is $q_u(\Theta - \theta - t)$. Therefore, the probability that no semi-qualified upload client is met in the $(t+1)$-th slot is $1 - \rho q_u(\Theta - \theta - t)$. The probability that no semi-qualified upload relay client is met during the entire upload search interval is thus $\prod_{t = 0}^{\Theta - \theta}(1 - \rho q(\Theta - \theta - t))$.

Similarly, let $\pi_d(s)$ be the fraction of clients whose last server meeting time was $s$ time slots ago. The invariant distribution can be solved according to another Markov chain as
\begin{align}
    \pi_d
    \begin{bmatrix}
    0 & 1 & 0 & ... & 0\\
    P'_\text{int}(1) & 0 & 1 - P'_\text{int}(1) & \cdots & 0\\
    P'_\text{int}(2) & 0 & 0 & \cdots & 0\\
    \vdots & \ddots & \ddots & \ddots & \vdots\\
    P'_\text{int}(\Delta) & \cdots & 0 & 0 & 1 - P'_\text{int}(\Delta)
    \end{bmatrix}
    = \pi_d
\end{align}
where $P'_\text{int}(t) = \frac{P_\text{int}(t)}{\sum_{s=t}^\Delta P_\text{int}(s)}$. Let $q_d(t) = \sum_{s = 0}^t \pi_d(s)$ be the cumulative distribution function. Suppose in the $t$-th slot in the download search interval, a client is met. The probability that this client is a semi-qualified download relay is $q_d(t)$. Therefore, the probability that no semi-qualified download client is met in the $t$-th slot is $1 - \rho q_d(t)$. The probability that no semi-qualified download relay client is met during the entire download search interval is thus $\prod_{t = 0}^{\Omega - \omega}(1 - \rho q_d(t))$. 

Finally, the probabilities of meeting at least one semi-qualified relay are obtained.

\section{Proof of Lemma \ref{lem:error_difference}}
\label{proof-l4}
Consider any time $t$, let $t'_i$ be the meeting time of client $i$ and its relay, by the definition of corrupted real sequence, we have
\begin{align}
    &\mathbb{E}\left[\|v^t - \widetilde{x}^t\|^2\right]
     = \mathbb{E}\left[\|v^t - x^t + x^t - \widetilde{x}^t\|^2\right]
    \nonumber\\\leq &  2\mathbb{E}\left[\|v^t - x^t\|^2\right] + 2 \mathbb{E}\left[\|x^t - \widetilde{x}^t\|^2\right]
      \nonumber\\ \leq & 2C^2(\theta, \Theta; \Delta)\eta^2 G^2 + 2\mathbb{E}\left[\|\frac{1}{N}\sum_{i \in U_t} \epsilon_i\|^2\right]
      \nonumber\\ \leq & 2C^2(\theta, \Theta; \Delta)\eta^2 G^2 + \frac{2}{N^2}\lvert U_t \rvert\sum_{i \in U_t}\mathbb{E}\left[\| \epsilon_i\|^2\right]
      \nonumber\\ \leq & 2C^2(\theta, \Theta; \Delta)\eta^2 G^2 + \frac{2q}{N^2}\lvert U_t \rvert\sum_{i \in U_t} \mathbb{E}[\|n^t_i \|^2]
      \nonumber\\ \leq & 2C^2(\theta, \Theta; \Delta)\eta^2 G^2 + \frac{2q}{N^2}\lvert U_t \rvert\sum_{i \in U_t} \mathbb{E}[\|\sum_{s= \tau^\text{last}_i(t)}^{t'_i-1}\eta g^s_i \|^2]
      \nonumber\\ \leq & 2C^2(\theta, \Theta; \Delta)\eta^2 G^2 + \frac{2q \Theta\eta^2 }{N^2}\lvert U_t \rvert\sum_{i \in U_t}\sum_{s= \tau^\text{last}_i(t)}^{t'_i-1} \mathbb{E}[\| g^s_i \|^2]
      \nonumber\\ \leq & 2C^2(\theta, \Theta; \Delta)\eta^2 G^2 + \frac{2q \Theta\eta^2 }{N^2}\lvert U_t \rvert\sum_{i \in U_t}\sum_{s= \tau^\text{last}_i(t)}^{t'_i-1} G^2
      \nonumber\\ \leq & 2C^2(\theta, \Theta; \Delta)\eta^2 G^2 + \frac{2q \Theta^2\eta^2 }{N^2}\lvert U_t \rvert^2 G^2
      \nonumber\\ \leq & 2C^2(\theta, \Theta; \Delta)\eta^2 G^2 + 2q \Theta^2\eta^2 G^2
\end{align}

Similar to the proof of Lemma \ref{lem:model_difference}, consider client $i$ at any time $t$ with the global model version $\psi_i(t)$. Therefore client $i$ is doing the local training steps using the global model $x^{\psi_i(t)}$ as the initial model. Thus, we use $x^{\psi_i(t)}$ as the anchor model to investigate the difference between $v^t$ and $x^t_i$. 
\begin{align}
    &\mathbb{E}\left[\left\|v^t - x^{t}_i\right\|^2\right] \nonumber\\=& \mathbb{E}\left[\left\|v^t - v^{\psi_i(t)} + v^{\psi_i(t)} - \widetilde{x}^{\psi_i(t)} + \widetilde{x}^{\psi_i(t)} - x^{t}_i\right\|^2\right]\nonumber\\
    \leq& 3\mathbb{E}\left[\left\|v^t - {v^{\psi_i(t)}}\right\|^2 + \left\|v^{\psi_i(t)} - \widetilde{x}^{\psi_i(t)}\right\|^2 + \left\|\widetilde{x}^{\psi_i(t)} - x^{t}_i\right\|^2\right]\nonumber\\
    \leq&3\mathbb{E}\left[\left\|\frac{1}{N}\sum_{i=1}^N \sum_{s= \psi_i(t)}^{t-1} \eta g^s_i\right\|^2\right] + 3\mathbb{E}\left[\left\|v^{\psi_i(t)} - \widetilde{x}^{\psi_i(t)}\right\|^2\right] \nonumber\\ &+ 3\mathbb{E}\left[\left\|x^{t'}_i - x^t_i\right\|^2\right]\nonumber \\
    \leq& 3\mathbb{E}\left[\left\|\frac{1}{N}\sum_{i=1}^N \sum_{s= \psi_i(t)}^{t-1} \eta g^s_i\right\|^2\right] + 3\mathbb{E}\left[\left\|v^{\psi_i(t)} - \widetilde{x}^{\psi_i(t)}\right\|^2\right] \nonumber\\&+ 3\mathbb{E}\left[\left\|x^{\psi_i(t)}_i - x^t_i\right\|^2\right]\nonumber\\
    \leq &3D^2(\omega, \Omega;\Delta)\eta^2 G^2 +6C^2(\theta, \Theta; \Delta)\eta^2 G^2 + 6q\Theta^2\eta^2G^2 \nonumber\\&+ 3{D^2(\omega, \Omega;\Delta)}\eta^2G^2\nonumber\\
    =& 6({D^2(\omega, \Omega;\Delta)} + C^2(\theta, \Theta; \Delta) + q\Theta^2)\eta^2 G^2
\end{align}

\end{document}